\documentclass[journal]{IEEEtran}
\IEEEoverridecommandlockouts                              % This command is only needed if 
\usepackage{graphicx} % for including images
\usepackage{subcaption} % for subfigures
\usepackage[export]{adjustbox}            % Adjust boxes in general and graphicx
\usepackage{multirow}
\usepackage{algorithm} 

\usepackage[noend]{algpseudocode}         % Algorithmix style like Algorithmic
\usepackage{amsmath}                      % Math macros and such
\usepackage{amssymb}                      % Math symbols
\usepackage{cite}                         % Better citation formatting  
\usepackage{graphics}                     % Image manipulation
\usepackage[normalem]{ulem} % for \sout (strikeout text)
\usepackage[usenames,dvipsnames,svgnames,x11names,table]{xcolor}
\usepackage{rotating}
\usepackage[most]{tcolorbox}
\usepackage{url}
\usepackage{hyperref}
\usepackage[normalem]{ulem}  % put in your preamble

\restylefloat{algorithm}

\newcommand{\algorithmicinput}{\textbf{Input:}}
\newcommand{\algorithmicoutput}{\textbf{Output:}}
\newcommand{\INPUT}{\item[\algorithmicinput]}
\newcommand{\OUTPUT}{\item[\algorithmicoutput]}
\newcommand{\CALL}[1]{\textsc{#1}}

\newcommand{\V}{\mathcal{V}}
\newcommand{\M}{\mathcal{M}}
\newcommand{\E}{\mathcal{E}}
\newcommand{\T}{\mathcal{T}}
\newcommand{\Sol}{\mathcal{S}}
\newcommand{\Constraint}{\mathcal{C}}
\newcommand{\HG}{\mathcal{H}}
\newcommand{\B}{\mathcal{B}}
\newcommand{\W}{\mathcal{W}}

\title{\Huge Lazy-DaSH: Lazy Approach for Hypergraph-based Multi-robot Task and Motion Planning}

\author{Seongwon Lee$^{1}$, James Motes$^{1}$, Isaac Ngui$^{1}$, Marco Morales$^{1,2}$, and Nancy M. Amato$^{1}$%
\thanks{$^{1}$Parasol Lab, University of Illinois Urbana-Champaign, USA
        {\tt \{sl148, jmotes2, ingui2, moralesa, namato\}@illinois.edu}}%
\thanks{$^{2}$ Department of Computer Science, Instituto Tecnológico Autónomo de México (ITAM),
            Mexico City, 01080, México. {\{\tt marco.morales\}@itam.mx.}}%
}

\begin{document}

\maketitle

\begin{abstract}
We introduce Lazy-DaSH, an improvement over the recent state of the art multi-robot task and motion planning method DaSH, which scales to more than twice the number of robots and objects while achieving an order of magnitude faster planning when applied to a multi-manipulator object rearrangement problem. We achieve this improvement through a hierarchical approach, where a high-level task planning layer identifies planning spaces required for task completion, and motion feasibility is validated lazily only within these spaces. In contrast, DaSH precomputes the motion feasibility of all possible actions, resulting in higher costs for constructing state space representations. 
{\color{black} Lazy-DaSH ensures efficient query performance by utilizing a hierarchical constraint feedback mechanism, effectively conveying motion feasibility to the query process while incrementally expanding the task and motion space representations when failures are detected, so the search space grows only as needed}. By maintaining smaller state space representations, our method significantly reduces both representation construction time and query time. We evaluate Lazy-DaSH in four scenarios, demonstrating its scalability with increasing numbers of robots and objects, as well as its adaptability in resolving conflicts through the constraint feedback.
\end{abstract}

\section{{Introduction}}\label{sec:introduction}
Multi-robot systems are used in domains such as warehouse operations and assembly, enabling faster completion through parallel operations and achieving more complex tasks through coordination.
Planning for such applications is challenging as the size of the planning space grows exponentially as both the number of robots and tasks increases and as the level of coordination required increases~\cite{c-crmp-88}.
When the coordination required is low, decoupled multi-robot motion planning (MRMP) addresses this complexity by decomposing the search space into independent robot state spaces and later resolves conflicts between individual robot plans.
However, multi-robot task and motion planning (MR-TMP) problems that require high levels of coordination are traditionally solved with coupled methods that directly consider the composite space of the system.
This enables the necessary coordination but suffers from the exponential scaling of the search space.
Recent work has explored hybrid approaches that aim to balance the strengths of coupled and decoupled approaches while minimizing their weaknesses.

\begin{figure}[t!]
    \centering
    \includegraphics[scale=0.32]{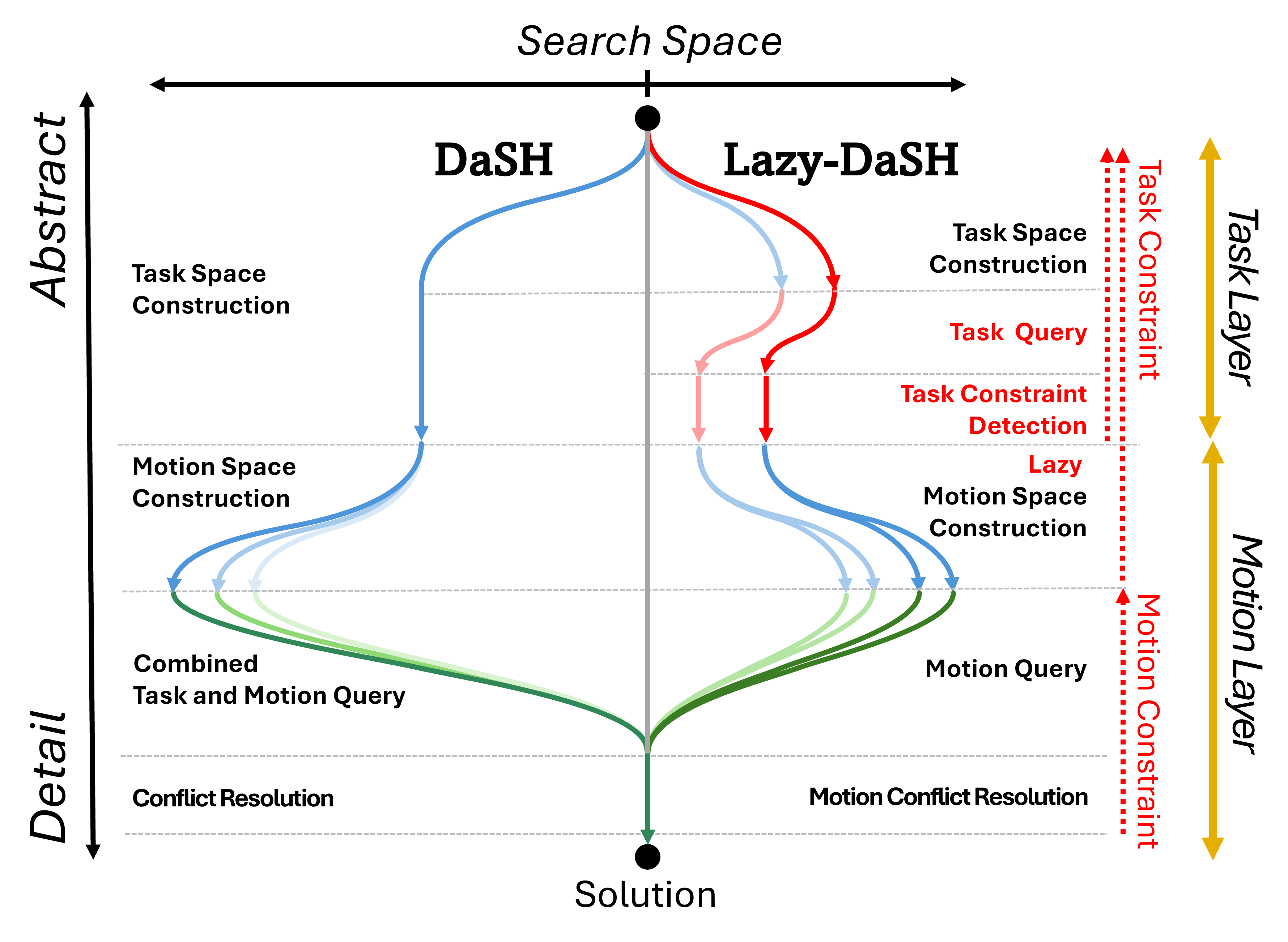}
    \caption{
    A comparison of the search space scope during the search processes of DaSH and Lazy-DaSH. 
    The introduction of the {\color{black}task space expansion}, task query, {\color{black}task constraint detection, task and motion constraint feedback,} and lazy motion validation phases distinguishes our approach from DaSH, as highlighted in red. 
    The task query narrows the search space, while lazy motion validation considers only motions in the candidate plan, reducing the computational cost of motion space construction. {\color{black}The task and motion constraint feedback initiates the expansion of the task space and motion space, thereby broadening the search space in the respective planning representations.}
    While both DaSH and Lazy-DaSH iteratively update representations upon plan failure, Lazy-DaSH employs a constraint feedback mechanism within a hierarchical framework to effectively manage both task-level and motion-level constraints, as illustrated in Fig.~\ref{fig:overview} and Algorithm~\ref{alg:lazy-dash}.
    }
    \label{fig:thumbnail}
\end{figure}

The \textbf{D}ecomposable St\textbf{a}te \textbf{S}pace \textbf{H}ypergraph (DaSH) framework \cite{motes2023hypergraph} is a hybrid approach for MR-TMP which seeks to focus computation effort only where coordination is needed through a more efficient hypergraph-based representation of the task space than traditional graph-based composite methods and produces smaller search spaces when considering the motion feasibility for groups of coupled robots and tasks.
The original DaSH framework computes the motion feasibility within each of these smaller search spaces and then uses this information within a combined task and motion planning query (Fig.~\ref{fig:thumbnail}).
This leads to fast query times, but the construction time required to compute representations can be excessive, as high levels of task and motion coordination are required with the growing number of robots and objects. 
{\color{black}In addition, queries that rely solely on motion-level representations often struggle in environments with task-oriented constraints.}

{\color{black}This paper presents Lazy-DaSH, which extends DaSH by managing representation size within a hierarchical structure.
Instead of exhaustively expanding the motion-level representation, Lazy-DaSH selectively expands both task- and motion-level representations under the guidance of task and motion constraint management.
This selective expansion lowers the cost of representation construction, while motion feasibility is evaluated lazily, only within the decoupled planning spaces required by the current task plan.
By deferring motion validation and using constraint management to refine the search space, Lazy-DaSH reduces the overhead of operations such as collision checking and focuses computational effort on sequencing and identifying critical tasks and motions.
Through the integration of efficient representation control, hierarchical constraint management, and lazy motion validation, Lazy-DaSH achieves a balance between computational efficiency and effective reasoning over task-oriented constraints during the query process.

Our contribution can be summarized as a new MR-TMP algorithm, Lazy-DaSH, which features:
\begin{itemize}
    \item A new task planning layer within the DaSH hierarchy, equipped with a queryable task representation and task query strategy.
    \item Lazy motion validation of robot state spaces included in the task plan, improving computational efficiency.
    \item Hierarchical constraint management for efficient control of representation size, expanding task and motion space representations only when necessary.
    \item Scalable and efficient replanning, handling more than twice the number of robots and objects compared to DaSH and achieving up to two orders of magnitude faster planning times in multi-manipulator object rearrangement scenarios.
\end{itemize}
}

\section{\color{black} Preliminaries and Related Work}\label{sec:related_work}
In this section, we introduce the preliminaries and related work for MR-TMP. 

\subsection{{Motion Planning }}
The position of a robot is completely parameterized by its degrees of freedom (DOF), which include its pose, orientation, and joint angles, and their values define a robot's configuration.
The set of all robot configurations forms the \textit{configuration space} ($\mathcal{C}_{space}$).
The motion planning problem seeks to find a continuous path between a start and goal configuration through the subset of valid configurations ($\mathcal{C}_{free}$).

Representing the high-dimensional $\mathcal{C}_{space}$ explicitly is intractable in the general case~\cite{c-crmp-88}.
To address this, sampling-based motion planning~\cite{kslo-prpp-96,l-rrtntpp-1998,bk-ppulp-00,kkl-aprpp-96,kl-rrtceaspp-00,orthey2023sampling} randomly samples configurations in $\mathcal{C}_{space}$ to construct a graph or tree representation that approximates connectivity.
Motion planning queries are answered by connecting the start and goal to this representation and computing a path over it. 

{\color{black} The main computational cost in this process comes from validating sampled configurations and edges, which requires repeated collision checking against the environment~\cite{lozano1983spatial,petrovic2018motion}.
To address this, methods such as lazy evaluation of edges~\cite{bohlin2000path,bk-ppulp-00}, reuse of previous collision-check results~\cite{hsu2006probabilistic}, adaptive or biased sampling strategies~\cite{gammell2014informed}, and hybrid search-plus-optimization planners ~\cite{ortiz2024idb,ortiz2024idb2} have been proposed to reduce unnecessary checks while still ensuring correctness.}

\subsection{Task and Motion Planning}
When object manipulations are required, the robot needs to navigate and interact with these objects to achieve task objectives.
This involves two layers: a task planning layer that determines the sequence of abstract actions and a motion planning layer that computes feasible motions to perform these actions.
Integrating these layers allows robots to satisfy the constraints imposed by the robots and the environment. 
% \subsubsection{{\color{black}Constraint Satisfaction Problem}}
There are three major categories of approaches that focus on constraint satisfaction \cite{garrett2021integrated}{\color{black}: \textit{sequencing-first, satisfaction-first, and interleaved approaches.}}

\textit{Sequencing-first} approaches plan high-level actions before determining the specific motions required to execute them \cite{srivastava2013using,garrett2018sampling,garrett2020pddlstream}. 
They assume all actions have feasible motions during the high-level planning, which is frequently not the case due to physical constraints. Thus, they include mechanisms for backtracking and trying alternative plans when initial ones fail.

\textit{Satisfaction-first} approaches focus on satisfying constraints related to continuous parameters (e.g., object positions and robot configurations) before creating an action sequence \cite{krontiris2015dealing,garrett2015ffrob,hl-mmmpnes-10}. 
They are efficient when it is easier to sample and test these parameters upfront rather than repeatedly attempting to fit them into an action sequence that might not work. Thus, they often involve a cycle of sampling and testing for feasibility check. 

\textit{Interleaved} approaches blend the sequencing of actions with the determination of parameter values, dynamically adapting both methods \cite{barry2013hierarchical,thomason2019unified,kim2020monte,kim2020learning}. 
They enable a flexible and efficient planning process by minimizing unnecessary backtracking and reducing the computational overhead associated with either fully sequential or satisfaction-first methods.

{\color{black}
In all of these methods, the key consideration is identifying which constraints must be captured and how they should be managed to establish valid plans.
The constraints that determine task or motion feasibility are often geometric or kinematic in nature and can be represented through task-space abstractions such as scene graphs~\cite{dantam2018incremental}, symbolic-geometric models~\cite{kaelbling2011hierarchical}, or constraint-based formulations~\cite{lagriffoul2014efficiently}.
By efficiently encoding these constraints across the hierarchical structure, planners can integrate motion-level feedback (e.g., collision checks or parameter failures) into task-level reasoning, thereby refining plans and guiding queries more effectively~\cite{garrett2021integrated,kaelbling2011hierarchical}.
}

{\color{black}

}

\subsection{Multi-robot Task and Motion Planning}

{\color{black} 
Multi-robot task and motion planning extends the fundamental concepts of task and motion planning to complex tasks that require collaboration between multiple robots.
This problem is significantly more complex when factoring in robot interactions in coordination, collision avoidance, and task allocation and scheduling \cite{motes2020multi}.
% The primary challenge is the combinatorial growth of the search space, and the three classes of approaches--coupled, decoupled, and hybrid--are applicable in this context as well. 
{\color{black}The primary challenge for multi-robot planning is maintaining the necessary coordination while accounting for the combinatorial growth of the search space. 
There are three standard approaches methods take to address this challenge: decoupled, which sacrifices coordination for smaller search spaces by planning for robots individually; coupled, which accepts the size of the search space in order to maintain coordination; and hybrid, which attempts to leverage the strengths of both coupled and decoupled while minimizing their weaknesses.}    

Most MR-TMP problems require a high level of coordination, and the decoupled approach often fails to find a feasible solution.
On the other hand, the coupled approach is capable of finding a highly coordinated solution but is not scalable due to the size of the search space. 
This creates a fundamental trade-off between solution optimality and computational scalability, which has driven the development of various planning strategies.

At one end of the MR-TMP spectrum, coupled approaches establish a benchmark for solution quality by aiming for theoretical guarantees like optimality. 
% Commonly, this is a major direction for Multi-robot Motion Planning and Multi-robot Pathfindng is problems.
% (Add contents here).
For example, the method proposed in \cite{hartmann2025benchmark} formalizes the multi-robot planning problem and provide asymptotically optimal baseline planners by operating in the full composite configuration space of all robots with multiple target goals. 
Such methods provide a foundational standard for asynchronous path quality but assumes the given task assignment.

To make planning tractable for large-scale, long-horizon applications, other approaches prioritize computational scalability \cite{mirrazavi2018unified}. 
Decoupled methods achieve this by planning for single or subgroup of robots individually and then coordinating their actions in a post-processing step to resolve conflicts. 
For example, \cite{hartmann2022long} presents breaking down a large problem into a series of smaller, more manageable subproblems to solve the long-horizon planning problem. 
The authors present a scalable planner for complex construction tasks by strategically decomposing the global problem into a sequence of smaller optimization problems. 
This approach enables the coordination of large, heterogeneous teams for tasks that are beyond the scope of fully coupled methods, trading theoretical optimality for the ability to solve highly complex, real-world scenarios.

Another key consideration for optimality and computational efficiency in MR-TMP is the execution model, which can be either synchronous or asynchronous. Synchronous approaches, where robots operate in lockstep, are often inefficient as faster robots are forced to wait for slower ones \cite{sb-smrgbmgwcc-20}.
Composite planning often enables asynchronousity by exposing partial orders over actions and scheduling them to compact the overall makespan, but computing globally optimal asynchronous schedules is typically intractable at scale \cite{huang2025apex}. 
To deal with the computational intractability of asynchronous planning, \cite{huang2025apex} first computes a Temporal Plan Graph (TPG), and post-process a completed plan into a more flexible partial-order representation, to make the system robust to real-world delays. 
In addition, asynchronous planning can be done by factoring time into the planner itself to create asynchronous motion among the robots involved in the tasks \cite{hartmann2022long,grothe2022st}.

Hybrid methods aim to balance the trade-offs between coupled and decoupled approaches. Many of these methods adapt the two-level structure of the grid world multi-agent pathfinding method, Conflict-Based Search (CBS), for the more complex TAMP domain.
For example, TMP-CBS \cite{motes2020multi} maps the CBS methodology to task planning to handle subtask dependencies. 
This approach decouples the problem at its low-level search, where it plans optimal paths for individual robots. 
The high-level search, however, operates in a coupled manner by identifying conflicts between these individual plans and imposing constraints to resolve them. 
While this approach demonstrated the effectiveness of hybrid methods in complex multi-robot task and pathfinding problems, it has not been validated for higher-dimensional planning scenarios, such as multi-manipulator task and motion planning problem.
DaSH \cite{motes2023hypergraph} employs a hypergraph representation to concisely model state spaces and queries plans in a decoupled manner and coupling tasks only when constraints require synchronization, while enabling asynchrounous planning. 
While powerful, its scalability is limited by its representation construction, which pre-computes motion feasibility for all possible actions. 
This high upfront cost becomes a bottleneck in scenarios with many robots, objects, and complex geometric constraints, as the number of potential actions grows exponentially.

\subsection{Multi-manipulator Object Rearrangement Problem}
Multi-manipulator object rearrangement is an important problem in MR-TMP requiring complex multi-robot coordination. 
Rather than focusing on symbolic Planning Domain Definition Language search \cite{haslum2019introduction,garrett2021integrated}, the multi-manipulator rearrangement problem emphasizes multi-modal reasoning over discrete modes (pick, place, handover, support) interleaved with continuous, collision-free motions \cite{motes2023hypergraph,hl-mmmpnes-10,hn-rmmpfahrmat-11,hl-itaprmmpdwmimpq-09}. 

Most research on the multi-manipulator object rearrangement problem has focused on developing efficient representations to encode the multi-modality using the coupled approach. 
For example, the authors in \cite{gao2022toward} utilize a shared manipulator workspace to create a shared space graph, enabling reasoning about multi-robot cooperation and adapting a path planning heuristic for multi-manipulator tasks. 
Moreover, a series of studies \cite{dk-praafpmam-15,shome2019anytime,shome2020drrt} has focused on developing a concise object mode graph \cite{hl-mmmpnes-10}. 
This graph captures valid transitions for pick, place, and hand-over actions and serves as a heuristic for guiding multi-modal motion planning \cite{hl-mmmpnes-10}. 
Building on the work in \cite{shome2019anytime}, \cite{sb-smrgbmgwcc-20} utilizes the object-centric mode graph for multiple objects and applies multi-agent pathfinding techniques to generate non-conflicting sequences of object modes.
For more complex assembly tasks, \cite{chen2022cooperative} employs a Mixed-Integer Linear Programming method for task assignment but did not effectively demonstrate scalability with respect to the number of robots, primarily due to the computational demands of roadmap generation and annotation processes.

Although representations that encode object–robot relations at both the task and motion levels provide a reasonable foundation for efficient querying, the composite state space often limits scalability when handling multiple robots and objects.  
It also tends to enforce synchronous planning, requiring robots to start and finish at the same global time.  
These restrictions reduce efficiency and frequently necessitate additional post-processing to yield a more compact task plan.  

% Another key consideration in MR-TMP is the execution model, which can be either synchronous or asynchronous. 
% Synchronous approaches, where robots execute actions in lockstep, simplify coordination but restrict concurrency and leave many robots idle \cite{sung2024asynchronous}. 
% In contrast, asynchronous execution allows robots to operate on independent timelines, enabling overlap of actions and generally reducing makespan \cite{cite}. 
% Although this adds complexity in scheduling and coordination, it expands the feasible solution space and better exploits parallelism.

DaSH \cite{motes2023hypergraph}, a recent hybrid and asynchronous MR-TMP method, employs a hypergraph representation to concisely model hybrid robot state spaces and has been validated in the multi-manipulator object rearrangement problem. 
It queries plans in a decoupled and parallel manner, while coupling decomposed tasks when constraints require synchronization to resolve conflicts among the decoupled tasks. 
While it shows performance improvements--three orders of magnitude faster than the benchmark presented in \cite{sb-smrgbmgwcc-20}--its scalability is limited, particularly in scenarios with complex constraints. 
For example, in tasks requiring geometric constraints or multiple steps for completion, DaSH cannot fully leverage task-inferred constraint information during the query process. 
This limitation arises because a \textit{combined} task and motion query in DaSH relies only on motion-collision constraints for replanning when the query fails, rather than incorporating task-specific constraints. 
In Section~\ref{sec:validation}, we demonstrate improved performance of Lazy-DaSH in environments with geometric constraints, even with a higher number of robots and objects.
}

\section{Problem Definition}
This section defines and explores the properties of the \textit{task space}, defining key terminology related to the MR-TMP problem we address in this paper and its application to the multi-manipulator object rearrangement problem.

\subsection{Task Space }
In the MR-TMP framework \cite{motes2023hypergraph}, the \textit{task space} $\T$ is defined by three key components: the movable bodies \(\B\), the configuration space \(\W\), and the constraints \(\mathcal{C}\). 
Each \textit{task space element} \(T_i = (B_i, W_i, C_i) \in \T\) consists of a subset of the movable bodies \(B_i \subseteq \B\), its associated configuration space \(W_i \subseteq \W\), and a set of task space constraints \(C_i \subseteq \mathcal{C}\) that define the valid of configurations space, \(W_i\), for the bodies $B_i$.
{\color{black}
Constraints encode the system's physical limitations (e.g., kinematics, collisions) and task requirements (e.g., stable grasps).}

An MR-TMP problem has a set of \textit{admissible} task space elements $\T^* \subseteq \T$, where the movable bodies and constraints of \(T_i \in \T^*\) define a valid set of configurations for the entire system.
The planner can consider \textit{transitions} between sets of task space elements.
These may reflect changes in the compositions of the subsets of moveable bodies and/or in the constraint sets included in elements.

\subsection{Multi-manipulator Object Rearrangement Problem}
In scenarios where manipulators are used to rearrange objects, the admissible task space elements $\mathcal{T}^* \subseteq \mathcal{T}_{\texttt{MANIP}}$ include robots, objects, and every possible combination of manipulator-object grasp pairs. 
These task space elements encompass not only the manipulators and objects themselves but also the manipulator-object pairs, along with their associated configuration spaces and constraints.
{\color{black}
We define three primary task space constraints. The \textit{stability constraint} $s_c(o_j,q)$ for an object $o_j$ is satisfied when the object is placed in a stable pose at $q$. 
The \textit{hand-free constraint} $h_c(r_i)$ for a robot $r_i$ is satisfied when the robot is empty-handed. 
The \textit{formation constraint} $f_c(r_i,o_j,t)$ for a robot–object pair $(r_i, o_j)$ is satisfied when the pair maintains a predefined stable grasp formation, where $t \in SE(3)$. 
Using these definitions, a task space element corresponding to a free object $o_j$ must satisfy $s_c(o_j,q)$, while an element representing a grasped pair $(r_i, o_j)$ must satisfy $f_c(r_i, o_j,t)$ for both the robot and object, with $h_c(r_i)$ no longer valid.
}

% {\color{orange}\textit{Transitions} (Old) are performed by pick, place, or hand-over actions.
% A pick action on an object, for example, removes the constraint that the object is on a stable surface and adds a constraint that the object is in a stable grasp by the robot while changing the composition from independent elements, one with a robot and one with an object, to a coupled robot-object composition in the post-transition element. 
% Transitions between task space elements are only possible at valid configurations between objects and robots for grasping. 
% }
{\color{black}
\textit{Transitions} between task space elements are realized through pick, place, or hand-over actions that explicitly modify the active constraint sets, thereby moving the system to new task space elements. 
Each action is characterized by preconditions and postconditions defining the resulting element(s). 
For example, a pick action on object $o_j$ by robot $r_i$ transitions the system from a state represented by two independent task space elements--one for the free robot and one for the stable object--into a composite element representing the pair $(r_i, o_j)$. 
The preconditions for this transition require that the object is in a stable placement at configuration $q$ (i.e., $s_c(o_j,q)$ holds) and that the robot can achieve a valid grasp configuration. 
In the resulting composite element {\color{black}(with both the robot and the object)}, the original $s_c(o_j,q)$ constraint is replaced by the $f_c(r_i, o_j,t)$ constraint, capturing the stable grasp.
}
Consequently, a solution to this problem involves a discrete sequence of task space elements, along with continuous motion paths within the configuration space of each task space element.

\begin{figure*}[t!]
    \centering
    \includegraphics[trim=0 0cm 0 0cm, clip, scale=0.195]{new_figures/overview10.png}
    \caption{Illustration of the hierarchical structure of the proposed Lazy-DaSH, showing two manipulators ($R_1$ and $R_2$) rearranging an object ($O_1$). The task query phase and the task constraint feedback scheme, which distinguish it from DaSH, are highlighted with red lines. This feature is also emphasized in Algorithm~\ref{alg:lazy-dash}. Each layer of the hierarchy is detailed in corresponding sections. {\color{black}Note that the different types of grasp modes are omitted from these figures to improve clarity of visualization and conceptual explanation.}
}
    \label{fig:overview}
\end{figure*}

\section{The Lazy-DaSH Method }\label{sec:method}
{\color{black}In this section, we first provide an overview of the Lazy-DaSH method contrasting it with the original framework (Fig. 2). This is followed by detailed explanations of each layer of the hierarchical approach and a discussion on the properties of the method.}

\subsection{\color{black}Overview}\label{sec:overview}
The overall structure of Lazy-DaSH is illustrated in Fig.~\ref{fig:overview}.
As in DaSH~\cite{motes2023hypergraph}, we adopt a hierarchical representation, using a hypergraph to model both the task space and the motion space.
In contrast to a standard graph, a (directed) hypergraph employs hyperedges (or hyperarcs) that connect multiple vertices simultaneously \cite{glpn-dhaa-93}.
{\color{black}Whereas a graph-based representation captures the composite state space, a hypergraph captures a hybrid state space, in which task space elements couple only when enforced by transition constraints.}
This property is particularly useful for compactly encoding inter-robot interactions and {\color{black}constraints on task space elements} within a relatively small representation compared to a standard graph \cite{motes2023hypergraph}.

{\color{black}Building on this hypergraph representation, both DaSH and Lazy-DaSH perform a representation-based query process.
DaSH integrates task and motion queries into a single process by performing the query on the motion-level representation, whereas Lazy-DaSH differentiates itself by adopting a two-stage hierarchy: a \textit{task query} in the task space hypergraph followed by a \textit{motion query} in the motion hypergraph.}
This creates a strict hierarchical structure with a task planning layer and motion planning layer, where each layer has its corresponding representation construction phase and its query phase.    
The representation construction phase encodes the satisfaction of task and motion constraints between robots and objects, while the query phase sequences the constraint-satisfied task and motion. 

Both Lazy-DaSH and DaSH perform hybrid and interleaved planning iterating between representation construction phases and query phases. 
However, they differ in their emphasis on sequencing and constraint satisfaction. 
DaSH prioritizes constraint satisfaction, as it pre-samples feasible configurations for each transition and generates feasible motions between each of these configurations before the query phase.
Conversely, to identify the minimum constraints required, Lazy-DaSH prioritizes sequencing, as the task query phase first sequences high-level actions before addressing constraint satisfaction. 
The construction and query phases for both approaches are interleaved, with constraints being fed back whenever a feasible solution cannot be found.

The planning process of Lazy-DaSH starts by entering the \textit{task planning layer}, first constructing the task space representation, or \textit{task space hypergraph} ($\HG_\T$), and subsequently generating a discrete task plan represented as \textit{task-extended hypergraph} ($\HG_{\texttt{TE}}$).
% {\color{orange} (Old) The resulting task plan is an \textit{unvalidated schedule} that presumes all motions within the task space elements are feasible and lazily validates the motion feasibility in the subsequent motion planning layer.}
{\color{black}The resulting task plan is an \textit{unvalidated schedule} that presumes all motions within the task space elements are feasible. 
The unvalidated schedule is first checked in the \textit{task conflict detection layer} against the static object poses and corresponding robot configurations in the {\color{black} unvalidated} history. 
Any invalid task sequences are then corrected by replanning with task constraints.
If a task query fails due to conflicts in task constraints, the task space representation is expanded to broaden the search space.
This approach reduces subsequent motion planning calls and collision checks in the conflict resolution layer by first identifying infeasible task sequences in advance. 
The feasibility of motions within the task plan is then lazily validated in the motion planning layer.}
The \textit{motion planning layer} first constructs the motion representation, or \textit{motion hypergraph} ($\HG_\M$), only for the elements of the task space identified as relevant by the task plan (unvalidated schedule). 
The motion feasibility is evaluated during the motion query phase by generating the \textit{motion-extended hypergraph} ($\HG_{\texttt{ME}}$).
This approach alleviates the need for an unnecessarily expensive motion feasibility check for actions not included in the task plan or occurring after an action with no valid motion.
The resulting motion plan is an \textit{optimistic schedule}, requiring the \textit{\color{black}motion conflict resolution layer} to generate the collision-free \textit{valid schedule}. 

Lazy-DaSH adopts a \textit{lazy} approach at three key stages within its hierarchical structure: first, during the task query phase, where it defers the motion feasibility check of the task sequence to the subsequent motion planning layer; and second, during the motion representation construction phase, where it assumes that motions are always feasible and postpones the feasibility check until the motion query phase; and lastly, during the conflict resolution layer where the lazy motion planning is used to replan the motion satisfying motion constraints.

The subsequent sections provide a detailed explanation of each layer in the hierarchy and a comprehensive comparison between the Lazy-DaSH and DaSH.

\begin{algorithm}[t!]
\small
  \caption{{\color{black}Lazy-DaSH Approach. }}
  \label{alg:lazy-dash}
  \begin{algorithmic}[1]
    \INPUT Initial task space element set $T_{\texttt{init}}$, goal task space element set $T_{\texttt{goal}}$, valid transition set $A$
    \OUTPUT Motion plan $\Sol_\M$
    \State $\Sol_\T, \Sol_\M \gets \emptyset$ \Comment{task and motion schedules}
    \State $\Constraint_\T, \Constraint_\M \gets \emptyset$\Comment{task and motion constraints}
    \While{$\Sol_\M$ not valid }
        \State {\color{blue} $\texttt{// Generate Task Space Hypergraph}$}
        \State $\HG_\T \gets \CALL{TaskSpaceHg}(\HG_\T, \Constraint_\T, A)$
        \State {\color{red} $\texttt{// Query Unvalidated Schedule}$}
        \State $\Sol_\T \gets \CALL{QueryTaskPlan}(\HG_\T,\Constraint_\T)$
        \State {\color{red} $\texttt{// Task-Vertex Conflict Check}$}
        \State $\Constraint_\T \gets \CALL{DetectTaskConflicts}(\Sol_\T)$
        \While{$\Sol_\T$ valid \textbf{and} $S_\M$ not valid}
            \State {\color{blue} $\texttt{// Generate Motion Hypergraph}$}
            \State $\HG_\M,\Constraint_\T,\Constraint_\M \gets \CALL{MotionHg}(\Sol_\T,\HG_\T,\HG_\M,\Constraint_\M)$
            \State {\color{ForestGreen} $\texttt{// Query Optimistic Schedule}$}
            \State $\Sol_\M,\Constraint_\T \gets \CALL{QueryMotionPlan}(\Sol_\T,\HG_\T,\HG_\M,\Constraint_\M)$
            \If{$\Constraint_\T \neq \emptyset$}
                \State \textbf{break}
            \EndIf
            \State {\color{ForestGreen} $\texttt{// Generate Valid Schedule}$}
            \State $\Sol_\M, \Constraint_\M \gets \CALL{MotionConflicts}(\Sol_\M,\Constraint_\M)$
        \EndWhile
    \EndWhile
    \State \Return $\Sol_\M$
  \end{algorithmic}
\end{algorithm}

\subsection{Task Planning Layer}\label{sec:task_plan_layer}
The highest layer of the hierarchy, the task planning layer, encompasses the construction of the task representation and its query phases within the task-level domain. 
% {\color{black} The representation encodes task spaces that abstract the planning entities and the feasible transitions between them. It begins with the minimal search space sufficient to capture the planning scene and is extended when the current task space representation is insufficient to find a solution.}
{\color{black} In an effort to keep the search space small, the initial set of object states only includes the start and goal position for each object. This is a greedy assumption that no intermediate object positions are required (which is the most common case). When that assumption fails, the method samples new object positions which satisfy the stability constraints and extends the representation to encode these object states.}

\subsubsection{Representation Construction (Task Space Hypergraph)}
The task planning layer captures the most abstract level of representations through the task space hypergraph $\HG_\T=(\V_\T,\E_\T)$, where each task vertex $v_\T=\langle T_i\rangle \in\V_\T$ represents a task space element $T_i\in \T$, and task hyperarcs $E_\T=\langle \texttt{Tail},\texttt{Head}\rangle \in\E_\T$ represent abstract transitions from the tail set comprising preconditions to the head set consisting of postconditions without explicit motion details.

For the multi-manipulator object rearrangement problems considered in this paper, the task vertices in $\HG_\T$ representing the task space elements correspond to robots by themselves {\color{black}($R_1$ and $R_2$ in Fig.~\ref{fig:overview})}, robots holding objects {\color{black}($R_1\!+\!O_1$ and $R_2\!+\!O_1$)}, or objects at a particular location as indicated by the constraints in the task space element {\color{black}($O_1^s$ and $O_1^g$)}. 
In Lazy-DaSH, the object-only task space elements differ from those in DaSH, where objects are associated with a more {\color{black}``location-specific" at configuration $q$ under constraints $s_c(O,q)$}. 
The location-specific constraints used here allow object-only task space elements to encode explicit object locations at the task space level {\color{black}($O_1^s$ and $O_1^g$ in Fig.~\ref{fig:overview}, representing the start and goal, respectively)}. 
This formulation enables $\HG_\T$ to be queried for generating a task plan (unvalidated schedule) that links start and goal object-only task space elements.

% {\color{black}The task space representation begins without additional object locations, keeping the task space minimal and avoiding an unnecessarily large search space. 
% When an extension is requested from the subsequent layer, specifically the task conflict detection layer, object-only task space elements are generated, and the task space representation is reconstructed to include these new elements while ensuring valid transition rules (e.g., grasp and handover) are satisfied.
% }

% {\color{orange}(Old) Generating $\HG_\T=(\V_\T,\E_\T)$ begins by connecting the set of allowed task space elements (robots, objects with locations, and robots with objects) with all possible transitions}
{\color{black}As in DaSH~\cite{motes2023hypergraph}, generating $\HG_\T=(\V_\T,\E_\T)$ begins with the initial task space elements (robot-only and object-only with stable poses) and recursively applies predefined transitions such as pick, place, and handover. 
Each transition couples or decouples task space elements according to their constraints, and the expansion continues until no additional feasible task space elements can be generated.}
The {\color{black}initial} object task space elements include the start (and goal vertex) for each object, which are all connected with a single hyperarc to task source vertex $v_\T^{\texttt{src}}$ (and task sink vertex $v_\T^{\texttt{sink}}$), to capture the full scope of the start (and goal) task states.
The reachability of objects in the start and goal locations is then evaluated based on the robot's capabilities (i.e., maximum payload and range) or grasp pose (i.e., a valid solution of inverse kinematics). 
{\color{black}The grasp pose may be evaluated across multiple candidates, such as sides, top, or other feasible configurations.}
However, the reachability only indicates that the object is within the graspable range of some robots, without considering the motion feasibility of the robot reaching the object.
Feasibility will be evaluated at the motion planning layer by running a motion planner.
% {\color{black}When the task space expansion is requested from the subsequent planning layer, indicating that the current task space repersentation is not containing enough search space to find the solution, additional object-only task space elements are introduced to the current task space representation and establish the connections based on the reachabilities, providing larger search space to explore.}
{\color{black}When the task space expansion extension is requested from the subsequent layer, specifically the task conflict detection layer, object-only task space elements are generated, and the $\HG_\T$ is reconstructed to include these new elements while ensuring valid transition rules (e.g., grasp and handover) are satisfied.}

The resulting $\HG_\T$ provides a \textit{queryable} task-level representation, encoding the abstract transitions between task space elements and reachability information for objects' start and goal states.

\subsubsection{Query (Task-extended Hypergraph)}\label{sec:task_query}
Since the hyperarcs in $\HG_\T$ transition between different task space compositions, directly querying the task plan within $\HG_\T$ may result in movable entities appearing in multiple task space compositions simultaneously. 
This may cause a task space element to perform multiple transitions at the same time, leading to an infeasible plan.
% {\color{orange} (Old) To avoid this issue, we generate the task-extended hypergraph $\HG_{\texttt{TE}}$, sequentially expanding the hyperarcs from the start task vertices by selectively choosing a viable transition for the frontiers in the current transition history.}
{\color{black}
To avoid this issue, we maintain a set of partial {\color{black}task hyperarcs, which represent potential transitions that are candidates for expansion but remain unexpandable until all of their tail task space elements are satisfied in the frontier of the current transition history.}
A hyperarc is expanded only if it does not result in a vertex with multiple outgoing hyperarcs.
{\color{black}
In DaSH \cite{motes2023hypergraph}, which performs integrated task and motion query, these partial hyperarcs are maintained within the motion hyperarcs, which enumerate more exhaustive options, including motions confined to the same task space elements as well as transitions for interacting with other robots and objects. 
In contrast, Lazy-DaSH preserves partial hyperarcs by abstracting motions and emphasizing task-oriented transitions.}
The expansion process is recorded by generating the task-extended hypergraph $\HG_{\texttt{TE}}$, where hyperarcs are sequentially expanded from the start task vertices by selectively choosing viable transitions from the current transition history. 
This enables independent task threads to be explored in parallel, supporting asynchronous execution.  

This method offers a distinct advantage in handling task constraints. 
If a potential transition is found to be infeasible, its corresponding partial hyperarc can be pruned without affecting other valid, parallel planning threads. 
Such flexibility is difficult to achieve in coupled planning approaches, which construct a single sequential plan and cannot easily modify independent branches of the search.    
}

The task-extended hypergraph is defined as $\HG_{\texttt{TE}}=(\V_{\texttt{TE}},\E_{\texttt{TE}})$.
Each task-extended vertex $v_{\texttt{\texttt{TE}}}=\langle v_\T,\Pi_{v_{\texttt{\texttt{TE}}}^{\texttt{\texttt{src}}}v_{\texttt{\texttt{TE}}}}\rangle$ $\in \V_{\texttt{TE}}$ is defined by a task vertex $v_\T \in \V_\T$ and a task-extended transition history $\Pi_{v_{\texttt{\texttt{TE}}}^{\texttt{\texttt{src}}}v_{\texttt{\texttt{TE}}}}$ that stores the history connecting from task-extended source vertex  $v_{\texttt{\texttt{TE}}}^{\texttt{\texttt{src}}}=\langle v_\T^\texttt{src},\emptyset\rangle$ to $v_{\texttt{\texttt{TE}}}$.
Each task-extended hyperarc $E_{\texttt{TE}}=\langle \texttt{Tail},\texttt{Head},E_\T\rangle \in \E_{\texttt{TE}}$ includes the information about the tail, head, and task hyperarc that contributes to the history transitions.
Beginning from $v_\texttt{TE}^{\texttt{src}}$, the search process finishes when the task-extended hyperarc finds the task-extended sink vertex $v_\texttt{TE}^{\texttt{sink}}=\langle v_\T^\texttt{sink},\Pi_{v_{\texttt{TE}}^{\texttt{\texttt{src}}}v_{\texttt{TE}}^{\texttt{\texttt{sink}}}}\rangle$ in a head set. 
The task plan is obtained by extracting the task vertices $v_{\texttt{TE}}.v_\T$ from each $v_{\texttt{TE}}$ along $\Pi_{v_{\texttt{TE}}^{\texttt{\texttt{src}}}v_{\texttt{TE}}^{\texttt{\texttt{sink}}}}$. The resulting task plan is an unvalidated schedule, which does not account for the motion feasibility along the task transitions.

{\color{black}In our formulation, task constraints are defined as $\Constraint_\T = \Constraint_h \cup \Constraint_f$, where $\Constraint_f = (v_\T^{\text{pre}}, v_\T^{\text{post}})$ denotes frontier constraints and $\Constraint_h = (v_\T^{\text{pre}}, v_\T^{\text{post}})$ denotes history constraints. A frontier constraint specifies that $v_\T^{\text{pre}}$ must not be present in the frontier for $v_\T^{\text{post}}$ to be expanded, while a history constraint specifies that $v_\T^{\text{pre}}$ must not appear in the history for $v_\T^{\text{post}}$ to be expanded. Frontier constraints tie hyperarc expansion to the \textit{current} state, whereas history constraints depend on the \textit{previous} states.

This formulation naturally captures geometric constraints: for example, an object placement in the current state may block the expansion of another task-space element, but once the object is moved, the expansion becomes feasible.
Moreover, the constraints are chainable, meaning a $v_\T^{\text{post}}$ can serve as a $v_\T^{\text{pre}}$ for another constraint.
We enforce these constraints by restricting hyperarc expansion, preventing the formation of partial hyperarcs whenever a prohibited vertex exists in the head set of the currently forming hyperarc.}

% {\color{black}In our problem setup, the task constraint is represented as $\Constraint_\T = (v_\T^{\text{prev}},v_\T^{\text{post}})$ that constrains order of the $v_\T$ in the task-extended history. This type of constriant can be identified in either the task-vertex conflict check phase (see \ref{sec:task_conflict_detection}) where object location history collides with other objects location or the robot's grasping configuration, or motion conflict resoultion phase (see \ref{sec:conflict_resolution_layer}) where recomputation of the robot's motion cannot solve the collision with placed objects or other robot's motions. The constraints prohibits the expansion of $E_\T$ while constructing the $\HG_{\texttt{TE}}$ if the $v_\T$ elements in $E_\T.\texttt{Head}$ trying to expands and $v_\T$ in $\Pi_{v_{\texttt{TE}}^{\texttt{\texttt{src}}}}$ can be matched with the pairs exsiting in the $\Constraint_\T$.} 
% {\color{black} TODO?: Task query algorithm with task constraints}

As discussed in \cite{motes2023hypergraph}, three query strategies are available for hypergraph-based planning: Dijkstra-like, A*-like, and greedy hyperpath queries.
In DaSH, the combined task and motion query explicitly computes motion costs, which are used to calculate admissible cost-to-go and hyperarc weights. 
This enables optimal solutions in both Dijkstra-like and A*-like hyperpath queries. 
By contrast, the greedy approach employs a heuristic to select the next hyperarc to add and backtracks if the selected actions fail to yield a feasible solution. 
It has been demonstrated that the greedy method achieves significantly faster and more scalable query performance while maintaining costs similar to those of optimal solutions and without compromising completeness \cite{motes2023hypergraph}.

In Lazy-DaSH, the task planning and motion planning layers are decoupled, requiring motion costs to be estimated rather than explicitly computed. This decoupled approach significantly reduces the computational effort required for representation generation, enabling faster and more scalable queries (Section~\ref{sec:analysis_and_discussion}).
Lazy-DaSH adopts a greedy search strategy guided by an estimated heuristic. For object rearrangement problems, an effective heuristic involves a distance-based approach: calculating the distance from the manipulator's base to the target object for grasping operations and the distance between the bases of two manipulators for handover operations \cite{sb-smrgbmgwcc-20}. 
We then compute an estimated cost-to-go from each vertex based on $\HG_\T$ and choose the action with the lowest estimated cost-to-go, backtracking when the actions are available.

\subsection{{\color{black}Task Conflict Detection Layer}}\label{sec:task_conflict_detection}
\begin{figure}[t]
    \centering
    \includegraphics[width=0.9\linewidth]{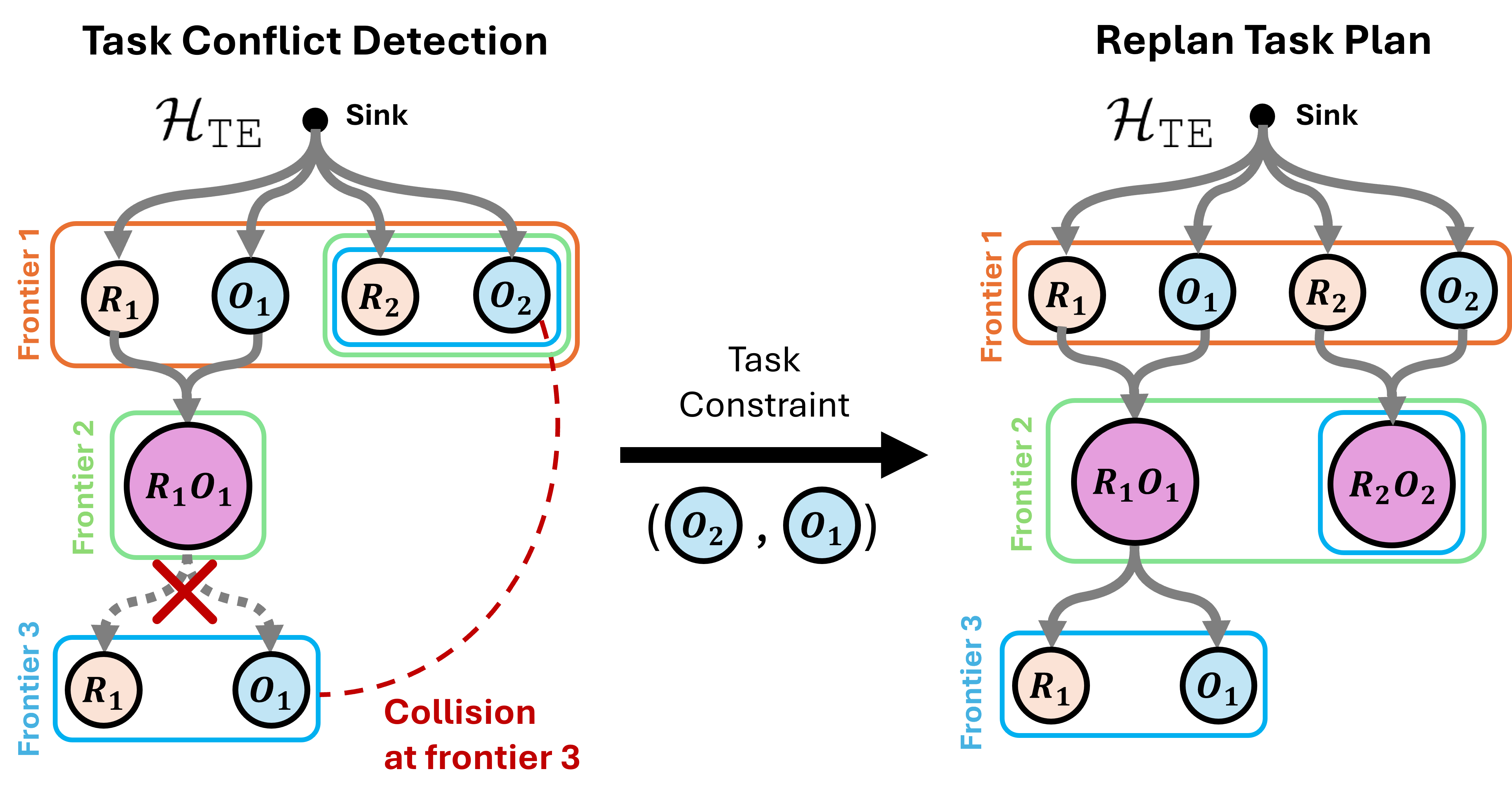}
    \caption{{\color{black}Illustration of task conflict detection. In the left figure, $O_1$ and $O_2$ collide at frontier 3, creating a task constraint that blocks the expansion of hyperarcs with $O_1$ at their head. Expansion of such hyperarcs becomes possible only when $O_2$ is absent at the frontier, in accordance with the task constraints.}}
    \label{fig:task_conflict_detection}
\end{figure}

{\color{black}
This section presents the second layer of the hierarchy, the task conflict detection layer, which validates the minimum feasibility of the unvalidated schedule by tracing transition configurations of robots and 
 {\color{black}static} objects throughout the schedule. 
{\color{black}An overview of this layer is provided in Algorithm~\ref{alg:task_conflict_detection} and Figure~\ref{fig:task_conflict_detection}.}

The unvalidated schedule assumes that all motions within the task plan are feasible, leaving the subsequent motion planning layer to confirm motion feasibility. 
However, relying solely on the motion planning layer to uncover all constraints (e.g., geometric constraints) becomes computationally expensive as the number of robots and objects increases.  

\subsubsection{Task Conflict}
To mitigate this, we identify task conflicts by sampling the minimum configuration information required to evaluate task-level feasibility. 
This includes the sequence of object poses across the schedule and the corresponding robot grasp configurations computed via inverse kinematics, guided by the formation constraint $f_c$ in the task vertex $v_\T$. 
These samples specify the explicit configurations $q$ at each transition point, which are then traced through the schedule to detect collisions. 
Detected conflicts are represented as motion vertex--vertex conflicts, where each motion vertex encodes the configuration of robots and objects at a transition in the unvalidated schedule. 
For example, such conflicts may arise when two objects occupy colliding positions, or when a robot in a grasping configuration collides with an unrelated object due to obstruction. 
{\color{black}Such conflicts are validated throughtout the entire unvalidated schedule to collect all conflicts existing in the current task plan, to entirely capture the potential task constraints.}

\subsubsection{Task Constraint Feedback}
{\color{black}When collisions involve objects that can be resampled (e.g., not anchored at start or goal poses), the conflict is resolved by generating alternative object poses within the planning area, which is handled directly in the task conflict detection layer.
In contrast, if a manipulator collides with an obstructing object, the resolution requires replanning by propagating task constraints back to the task query phase.
The identified constraints captures the entirety of the conflicts in the current unvalidated schedule and can be reused throught the task query phase afterwards. 
}

{\color{black}However, task constraints may conflict when the order of task space vertices is reversed, leading to contradictions.
To resolve this, additional stable poses must be introduced as move-out states, requiring new task space elements, since the minimal initial representation may be insufficient to satisfy all detected constraints during the task query process.
In such cases, the task conflict detection layer identifies the object involved in the unsatisfiable constraints and expands the task space representation with an additional stable pose, thereby expanding the hypergraph to include the new transitions.}

The iterative loop of querying, detecting task constraints, {\color{black}and expanding the task space} continues until a task-level feasible plan is generated, which is then forwarded to the motion planning layer for further validation.
}

\begin{algorithm}[t]
\caption{Task Conflict Detection}
\label{alg:task_conflict_detection}
\begin{algorithmic}[1]
\Require Unvalidated schedule $\Pi_{v_{\texttt{TE}}^{\texttt{\texttt{src}}}v_{\texttt{TE}}^{\texttt{\texttt{sink}}}}$, task-space hypergraph $\HG_\T$
\Ensure Task constraint set $\Constraint_\T$
\State $\Constraint_\T \gets \emptyset$
\State $f \gets \{\,v_\T^{\texttt{src}}\,\}$ \Comment{frontier initialized at source}
\For $E_\texttt{TE} \in \Pi_{v_{\texttt{TE}}^{\texttt{\texttt{src}}}v_{\texttt{TE}}^{\texttt{\texttt{sink}}}}$
  \State $q \gets \textsc{GetTransitionConfigs}(f, E_\texttt{TE}.E_\T)$
  \State $C \gets \textsc{CollisionCheck}(f, q)$
  \If{$C \neq \emptyset$}
    \State $\Constraint_\T \gets \Constraint_\T \cup C$
  \EndIf
  \State $\textsc{ApplyAction}(f, E_\texttt{TE}.E_\T.\texttt{Tail})$
\EndFor
\State \Return $\Constraint_\T$
\end{algorithmic}
\end{algorithm}

\subsection{Motion Planning Layer}\label{sec:motion_plan_layer}
This section describes the {\color{black}third} layer of the hierarchy, the motion planning layer, which involves the construction of the motion representation and the query phases within the motion-level domain.
\subsubsection{ Representation Construction (Motion Hypergraph)}
The motion hypergraph $\HG_\M=(\V_\M,\E_\M)$ captures the motion details of $\HG_\T$.

{\color{black}
Each motion vertex $v_\M = \langle v_\T, q \rangle \in \V_\M$ consists of a task-space vertex $v_\T$ and an explicit configuration $q$ sampled either by inverse kinematics or by the motion planner. 
This representation links configurations with the associated bodies, configuration spaces, and constraints defined in the task-space vertices. 
Each motion hyperarc $E_\M = \langle \texttt{Tail}, \texttt{Head}, E_\T, \pi \rangle \in \E_\M$ is defined by its tail and head sets, the corresponding task hyperarc $E_\T$, and a configuration path $\pi$ that connects the motion vertices in $E_\M.\texttt{Tail}$ to those in $E_\M.\texttt{Head}$.
As in DaSH~\cite{motes2023hypergraph}, the motion hyperarcs can be categorized as a {\color{black}composition} $E_\M^{\texttt{comp}}$, transition $E_\M^{\texttt{tran}}$, or move hyperarc $E_\M^{\texttt{move}}$. 
In our problem definition, $E_\M^{\texttt{tran}}$ denotes grasping or handover motions. 
For example, in a pick transition, the tail configurations include the object’s pose and the robot’s grasp configuration, each associated with a task-space vertex. 
The head configuration is the composite state of the robot grasping the object, where the grasp pose is constrained by the formation constraint $f_c$ in the task-space vertex $v_\T$. 
Any motion involving either a empty-handed robot or a robot holding an object is represented as $E_\M^{\texttt{move}}$.

% The transition sampler determines transition configurations (e.g., grasp and hand-over pose computed by inverse kinematics) only for transitions between the relevant task space elements. 
{\color{black}Since the configurations at transition points (e.g., robot’s grasp pose) are already sampled and verified in the task conflict detection layer to provide the minimum information needed for task plan validation, these configurations directly specify the tail and head sets of $E_\M^{\texttt{tran}}$.}
% The transition configurations are verified each other along the unvalidated schedule, ensuring there is not conflict between transition configurations in orders.  
This creates anchor points that the motion query must visit, bridging the $E_\M^{\texttt{move}}$ generated from different task space compositions. 
Roadmaps are then generated to serve as the basis for planning both $E_\M^{\texttt{tran}}$ and $E_\M^{\texttt{move}}$ during the subsequent motion query phase. 
In the Lazy-DaSH framework, these roadmaps are generated in an edge-unvalidated form {\color{black}following the principles of Lazy-PRM~\cite{bohlin2000path}}, with the feasibility of each motion deferred to validation during the subsequent query phase.

However, transition configurations may turn out to be invalid if they result in collisions with the environment, in which case resampling is performed to search for valid alternatives. 
{\color{black}For example, the transition configurations for a handover operation can be resampled by considering alternative mid-air poses of the object to avoid collisions with the environment.}
If the resampling attempts fail within the given budget, meaning no collision-free configuration exists to compose the $E_\M^{\texttt{tran}}$, the transition is considered infeasible. 
The planner then introduces task constraints that prohibit the corresponding task sequence, thereby preventing the expansion of the task hyperarc $E_\T$ and excluding the infeasible interaction during task plan replanning.
}
\subsubsection{Query (Motion-extended Hypergraph)}
% {\color{orange} (Old) Similar to $\HG_{\texttt{TE}}$, the motion sequence is queried by the motion-extended hypergraph $\HG_{\texttt{ME}}$ to avoid movable entities attempting multiple motions simultaneously.}
Since the abstract guideline of the motion plan is defined by the task plan, the motion-extended hypergraph is basically a motion-detailed version of the task plan.  
The feasibility of the motions between these transition configurations is validated while tracing the task plan.
{\color{black}
The motion feasibility of task plan reflects the feasibility of $E_\M.\pi$ along the unvalidated schedule and is sequentially validated in a lazy manner by querying a lazy sampling-based motion planner such as Lazy-PRM \cite{bk-ppulp-00}.
The lazy motion validation approach minimizes unnecessary computational effort by deferring validation until it becomes necessary. 
If a motion query fails while tracing the unvalidated schedule, local roadmap improvements are attempted on the edges where coarse motion validation has succeeded within the given budget (e.g., timeout).
If these improvements fail, additional vertices are sampled globally in the roadmap to search for an alternative global path.
However, if motion planning still fails after a certain amount of effort (e.g., exceeding the timeout or the maximum number of roadmap improvement iterations), the process falls back to the task query phase, which backtracks to explore an alternative unvalidated schedule.
Importantly, the invalid motion hyperarc is not permanently excluded from the motion hypergraph, allowing it to be revisited if it appears again under a different task plan.
% This persistence ensures that feasible solutions are never prematurely ruled out, preserving the probabilistic completeness of the overall algorithm.
% If an invalid motion is found in the unvalidated schedule, it implies that the validity of the subsequent motions does not need to be checked yet as the task plan may change.
% the task constraints are provided to the task query phase to enforce restrictions on the unsuccessful task order, prompting a replanning of the task plan.
}

% {\color{orange}(Old) If motion validation fails after a certain amount of effort (e.g., timeout), the task constraints are provided to the task query phase to enforce restrictions on the unsuccessful task sequence, prompting a replanning of the task plan.}

The motion-extended hypergraph is defined as $\HG_{\texttt{ME}}=(\V_{\texttt{ME}},\E_{\texttt{ME}})$, similar to $\HG_{\texttt{TE}}$ but with motion information. 
Each motion-extended vertex $v_{\texttt{\texttt{ME}}}=\langle v_\M,\Pi_{v_{\texttt{\texttt{ME}}}^{\texttt{\texttt{src}}}v_{\texttt{\texttt{ME}}}}\rangle  \in \V_{\texttt{ME}}$ consists of a motion vertex $v_\M \in \V_\M$ and a motion-extended transition history $\Pi_{v_{\texttt{\texttt{ME}}}^{\texttt{\texttt{src}}}v_{\texttt{\texttt{ME}}}}$ that stores a history extending from the motion-extended source vertex $v_{\texttt{\texttt{ME}}}^{\texttt{\texttt{src}}}=\langle v_\M,\emptyset\rangle$ to $v_{\texttt{\texttt{ME}}}$.
Each motion-extended hyperarc $E_{\texttt{ME}}=\langle \texttt{Tail},\texttt{Head},E_\M\rangle \in \E_{\texttt{ME}}$ includes the information about the tail, head, and motion hyperarc that contributes to the history transitions.
The motion query process begins at $v_{\texttt{\texttt{ME}}}^{\texttt{\texttt{src}}}$, expands the motion hyperarcs guided by the task plan, and terminates at $v_{\texttt{\texttt{ME}}}^{\texttt{\texttt{sink}}}$. 

{\color{black} When querying motions along the unvalidated schedule, motion constraints are applied to prevent traversal through regions defined by the boundaries of colliding objects and robots.
These regions are identified in the subsequent layer, specifically motion conflict resolution layer, and incorporated into replanning requests.
The motion constraint is defined as $\Constraint_\M=(\Pi_{v_{\texttt{TE}}^{\texttt{src}}v_{\texttt{TE}}^{\texttt{1}}},\Pi_{v_{\texttt{TE}}^{\texttt{src}}v_{\texttt{TE}}^{\texttt{2}}},p)$, which consists of the motion histories leading to the colliding hyperarcs $E_{\texttt{TE}}^1$ and $E_{\texttt{TE}}^2$, together with $p$, a collision region in the task space derived from the geometries of the colliding objects.
Since collisions are history-dependent, we preserve the transition histories of colliding hyperarcs and use their last elements to identify the conflicting motions.
This allows the planner to mark regions for avoidance in context, enabling replanning without discarding the entire plan.
} 

The resulting motion history $\Pi_{v_{\texttt{ME}}^{\texttt{\texttt{src}}}v_{\texttt{ME}}^{\texttt{\texttt{sink}}}}$ in $v_{\texttt{\texttt{ME}}}^{\texttt{\texttt{sink}}}$ provides a motion plan that represents an optimistic schedule. 
This schedule ensures collision-free paths within each composition space but does not consider potential collisions with other moving bodies. 
The conflict resolution layer resolves these issues by refining individual motion plans to generate a collision-free schedule for all moving entities.

\subsection{{\color{black} Motion Conflict Resolution Layer}}\label{sec:conflict_resolution_layer}
This section presents the final layer of the hierarchy, the conflict resolution layer, designed to identify motion conflicts between robots or objects that were not considered as coupled state spaces during the generation of the optimistic schedule. 
Any unsolvable motion conflicts lead to the creation of motion/task constraints, intended to enforce restrictions for replanning the motion/task plan.

% \subsubsection{Conflicts}
% {\color{orange} (Old) There are three types of motion conflicts, arising from interactions between two motion vertices, two motion hyperarcs, or a motion vertex and a motion hyperarc. 
% A \textit{vertex-vertex} conflict occurs when two objects are in a colliding position and is often handled by resampling the object's positions within the planning area.
% A \textit{hyperarc-hyperarc} conflict involves a collision between two moving entities and is generally resolved by adjusting the timing of the motion using a motion scheduling algorithm, or by replanning to ensure the movable bodies avoid each other.
% A \textit{hyperarc-vertex} conflict can arise when a static entity obstructs the moving entity and is addressed by replanning motions to navigate around the obstruction or by reordering the task sequence for colliding entities to pass through in a coordinated order.
% It is worth noticing that motion conflicts can be detected both in the motion query phase in the motion planning layer and the conflict resolution layer (Fig.~\ref{fig:overview}). }
\subsubsection{Motion Conflicts}
{\color{black} There are two types of conflicts in the motion conflict resolution layer, arising from interactions between two motion hyperarcs, or between a motion vertex and a motion hyperarc. 
Note that conflicts between two motion vertices are already identified in the task conflict detection phase and resolved by replanning the task plan in the task planning layer (Section~\ref{sec:task_plan_layer}). 

A \textit{hyperarc--hyperarc} conflict involves a collision between two moving entities and is resolved by adjusting the timing of the motion using a motion scheduling algorithm, or by replanning with path constraints to ensure the moving bodies avoid each other. A \textit{hyperarc--vertex} conflict arises when a static object obstructs a moving entity and is addressed either by replanning the motion with constraints to navigate around the obstruction using motion constraints ($\Constraint_\M$), or by reordering the task sequence so that the colliding entities pass through in a coordinated order using task constraints ($\Constraint_\T$), and it is further described in the subsequent section.}

% \subsubsection{Motion Constraint Feedback}
% {\color{orange} (Old)
% An overview of the constraint feedback is illustrated in Fig.~\ref{fig:overview}.
% The hierarchical structure, comprising the task planning layer and the motion planning layer, enables the resolution of conflicts within these two distinct levels.
% The constraints are applied backward through the hierarchy.
% The motion conflicts are first addressed at the motion planning level by replanning the conflicting motions to find alternative motions. 
% If these attempts consistently fail (e.g., timeout), it is assumed that the current task plan cannot be executed with a feasible motion, leading to the generation of task constraints. 
% These task constraints are designed to produce an alternative task sequence that seeks a feasible motion plan.}

\subsubsection{Motion Conflict Resolution and Constraint Feedback}
{\color{black}
Motion conflicts can be detected and addressed in both the motion query phase of the motion planning layer and the motion conflict resolution layer (Fig.~\ref{fig:overview}), but the replanning strategies differ. 
In the motion query phase, failures are handled by searching for alternative motions or refining the roadmap. 
In contrast, the motion conflict resolution layer formulates a subproblem that defines a new composite space, enabling replanning in a larger motion search space. 
This approach is inspired by the Adaptive Robot Coordination (ARC) method~\cite{solis2024adaptive} and adapted to the task and motion planning framework, where task dependencies coordinate the initiation and termination of motions to ensure alignment with the timing of related tasks, as shown in Algorithm~\ref{alg:SARC} and Figure~\ref{fig:SARC}.

\begin{algorithm}[t]
\caption{Scheduled Adaptive Robot Coordination}\label{alg:optimistic_schedule_repair}
\label{alg:SARC}
\begin{algorithmic}[1]
\Require Task space hypergraph $\HG_\T$, motion hypergraph $\HG_\M$, task transition history $\Pi_{v_{\texttt{TE}}^{\texttt{\texttt{src}}}v_{\texttt{TE}}^{\texttt{\texttt{sink}}}}$, motion transition history $\Pi_{v_{\texttt{ME}}^{\texttt{\texttt{src}}}v_{\texttt{ME}}^{\texttt{\texttt{sink}}}}$.
\State $\Constraint_\M \gets \emptyset$
\State {\color{blue} $\texttt{// Build Dependency Graph}$}
\State $D \gets \textsc{DependencyGraph}(\Pi_{v_{\texttt{TE}}^{\texttt{src}}v_{\texttt{TE}}^{\texttt{sink}}})$
\State {\color{blue} $\texttt{// Collect Mapping $E_\T \rightarrow E_\M$}$}
\State $M \gets \textsc{MapTaskMotion}(\Pi_{v_{\texttt{TE}}^{\texttt{src}}v_{\texttt{TE}}^{\texttt{sink}}},\Pi_{v_{\texttt{ME}}^{\texttt{src}}v_{\texttt{ME}}^{\texttt{sink}}},D)$
\State {\color{blue} $\texttt{// Find Motion Conflicts}$}
\State $C_\M \gets \textsc{FindMotionConflict}(M,D)$
% \State $D \gets \textsc{BuildDependencyGraph}()$
\While{$C_\M \neq \emptyset$}
    \State $\mathcal{Q}, G \gets \textsc{CreateSubProblem}(C_\M,M,D)$
    \State $\E_\M \gets \textsc{SolveSubProblem}(M, \mathcal{Q})$
    \If{$\E_\M \neq \emptyset$} \Comment{conflict resolved}
        \State $\textsc{UpdateMotionHypergraph}(\HG_\M, \E_\M')$
        \State $C_\M \gets \textsc{FindMotionConflict}(M,D)$
    \Else
        \State \Return $\HG_\M, \Constraint_\M$
    \EndIf
\EndWhile
\end{algorithmic}
\end{algorithm}

\begin{figure}
    \centering
    \includegraphics[width=0.7\linewidth]{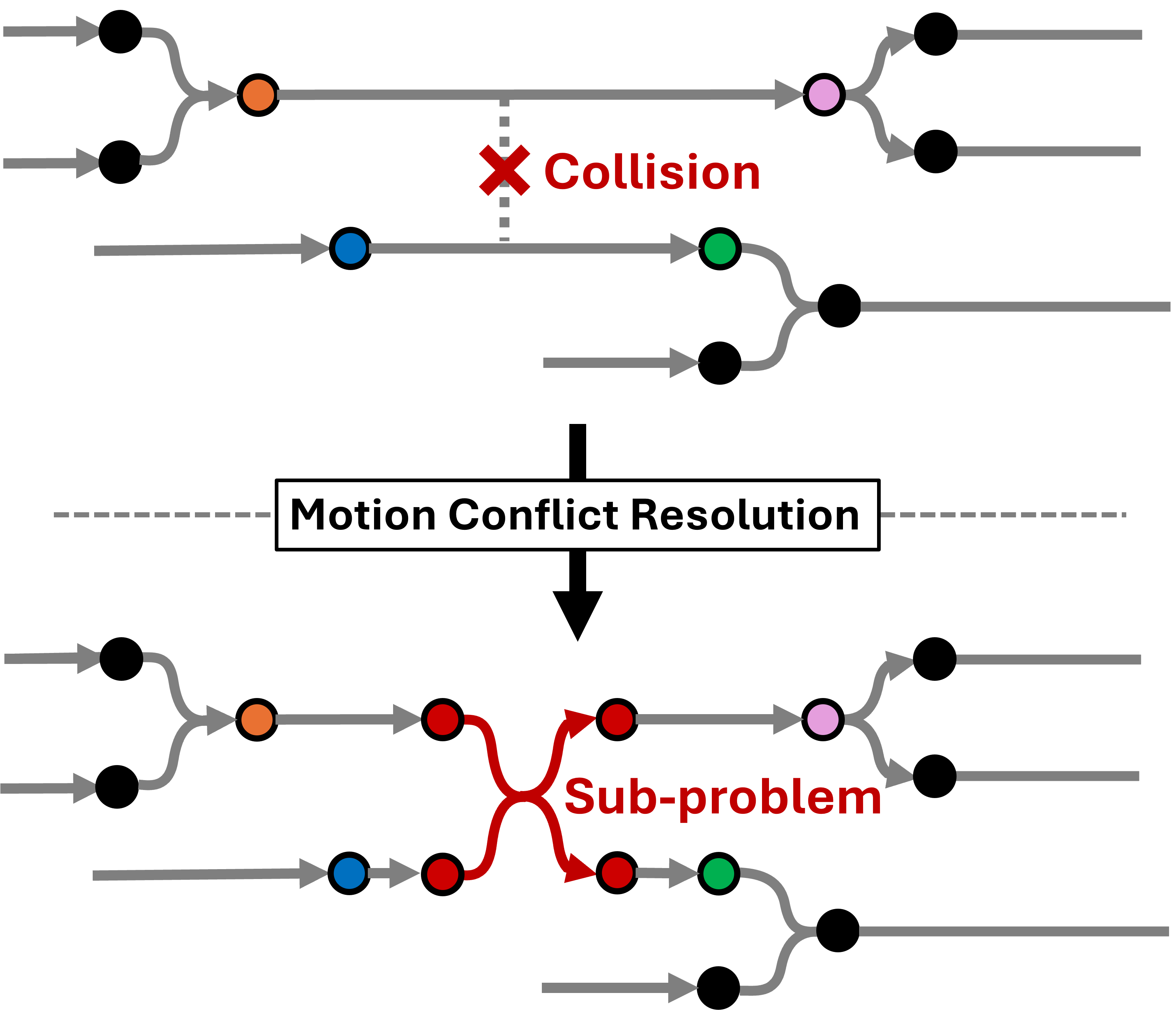}
    \caption{\color{black}Illustration of scheduled adaptive robot coordination. 
    }
    \label{fig:SARC}
\end{figure}

}

\subsection{Discussion }\label{sec:method_discussion}
{\color{black}This section compares Lazy-DaSH to DaSH by examining their emphasis on sequencing versus constraint satisfaction, their approaches to managing state space representations, and provides a discussion of probabilistic completeness.
}
\subsubsection{{\color{black}Comparison with DaSH}}
As discussed at the beginning of Section~\ref{sec:overview}, both Lazy-DaSH and DaSH can be categorized as hybrid and interleaved planning but differ in their emphasis on sequencing and constraint satisfaction, where DaSH prioritizes constraint satisfaction and Lazy-DaSH prioritizes sequencing.  
The approach for Lazy-DaSH aims to alleviate the exponential growth in representation size by identifying the minimum constraints required.

Comparing how DaSH and Lazy-DaSH manage their state space representations effectively illustrates the differences between the two approaches. 
From the perspective of the search process, the search space expands during the task/motion representation construction phases (blue arrows in Fig.~\ref{fig:thumbnail}) and is queried by the task/motion query phase (green and red arrows in Fig.~\ref{fig:thumbnail}).
DaSH constructs the representations in two immediate sequential phases, exhaustively adding every feasible transition and motion for each robot, object, and their interactions into the representation. 
This approach leads to a continuous expansion of search space across the representation construction layers (blue arrows). 
Then the combined task and motion planning query phase finds a solution within the constructed representation (green arrows). 
{\color{black}Conversely, Lazy-DaSH in Fig.~\ref{fig:thumbnail} narrows the search space by invoking the task query phase (red arrows in Task Query) immediately after constructing the task space representation (blue and red arrows in Task Space Construction). Furthermore, the task conflict detection layer postpones motion space expansion until a valid task plan is found (red arrows in Task Conflict Detection) and expands the task space elements only when necessary.}
This occurs before further expanding search space in the subsequent motion representation construction phase (blue arrows in Lazy Motion Space Construction phase).
This approach results in a more compact search space compared to DaSH, containing only the essential information, which facilitates faster query processes.

\subsubsection{{\color{black}Theoretical Properties}}
\label{sec:completeness}
{\color{black}

This section elaborates on the probabilistic completeness of Lazy-DaSH.
The property arises from its hierarchical, iterative structure, where representation construction and constraint-driven expansion ensure that the search space grows whenever the current representation is insufficient.

The guarantee builds on the properties of the task planning and motion planning components.
For a given task-extended hypergraph $\HG_{\texttt{TE}}$, a depth-first search with backtracking is complete and will always find a task plan if one exists within the representation $\HG_\T$.
For the motion layer, Lazy-PRM is probabilistically complete, so if a feasible motion exists, the probability of finding it approaches one as the roadmap becomes denser, with unvalidated edges eventually being evaluated.

To extend these guarantees to the combined planner, Lazy-DaSH relies on systematic constraint handling and representation expansion.
Lazy-DaSH treats each planning failure as a signal that the current representation does not fully cover the search space.
Feedback $(\Constraint_\T \text{ or } \Constraint_\M)$ drives expansion by adding new task vertices or hyperarcs to $\HG_\T$ and new motion vertices or hyperarcs to $\HG_\M$.
Task-level inconsistencies prompt $\HG_\T$ to be extended with new object placements or robot--object elements, while motion-level failures refine $\HG_\M$.
Importantly, constraints introduced from failures are scoped to the specific context ($\Pi_{v_{\texttt{TE}}^{\texttt{src}}v_{\texttt{TE}}}$ or $\Pi_{v_{\texttt{ME}}^{\texttt{src}}v_{\texttt{ME}}}$) that failed.  
For example, task constraints ($\Constraint_\T$) such as frontier constraints ($\Constraint_f$) and history constraints ($\Constraint_h$) are tied to the context of the current task query branch ($v_{\texttt{TE}}$) by considering frontiers ($v_\T$) or task transition history ($\Pi_{v_{\texttt{TE}}^{\texttt{src}}v_{\texttt{TE}}}$), while motion constraints ($\Constraint_\M$) are tied to the motion history ($\Pi_{v_{\texttt{ME}}^{\texttt{src}}v_{\texttt{ME}}}$) of colliding hyperarcs ($E_\M$).  
These constraints can be lifted when new task or motion samples are introduced, ensuring that feasible plans are never permanently excluded.

The overall process follows a consistent cycle in which the planner queries on the current $(\HG_\T,\HG_\M)$, detects failure, generates constraints, expands the representations, and then re-queries.
Two conditions ensure that this loop provides probabilistic completeness.
First, expansions are monotone, meaning they only add task and motion elements and never permanently remove a potentially feasible plan.
Second, expansions are exhaustive, meaning that as iteration proceeds, every feasible task sequence or motion configuration has a nonzero probability of being generated, and with increasing budgets the probability that all relevant interactions are eventually attempted approaches one.

Because query failures always trigger constructive expansion, Lazy-DaSH systematically enlarges its search space until it contains a feasible solution.
Therefore, under the standard assumptions for Lazy-PRM together with monotone and exhaustive expansion of $\HG_\T$ and $\HG_\M$, Lazy-DaSH is probabilistically complete.
}
\begin{figure*}[ht!]
    \centering
    \begin{subfigure}{0.24\textwidth}
        \centering
        \includegraphics[width=0.7\linewidth]{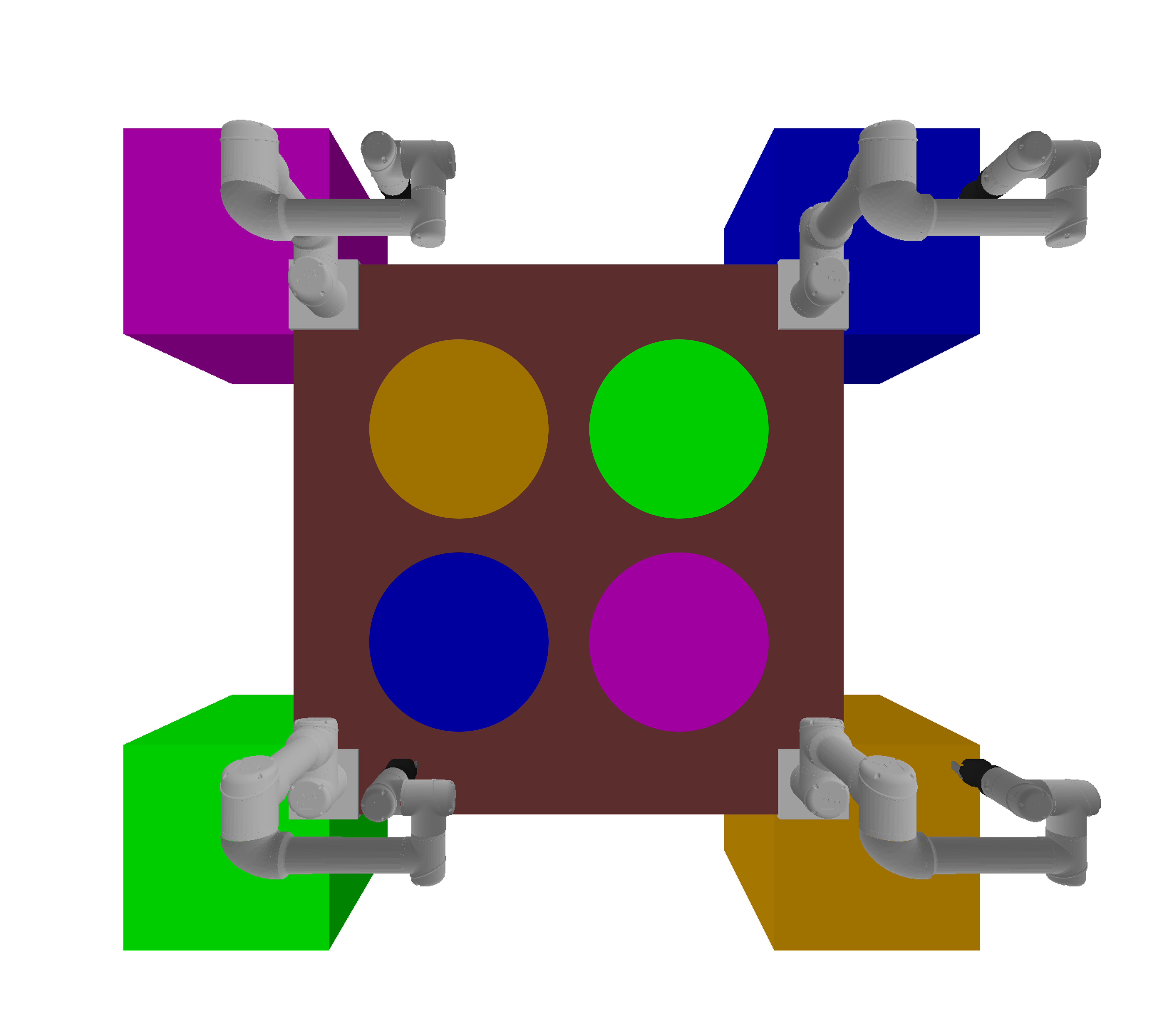}
        \caption{Sorting: 4 robots}
        \label{fig:4sort}
    \end{subfigure}
     \begin{subfigure}{0.26\textwidth}
        \centering
        \includegraphics[angle=0, width=0.5\linewidth]{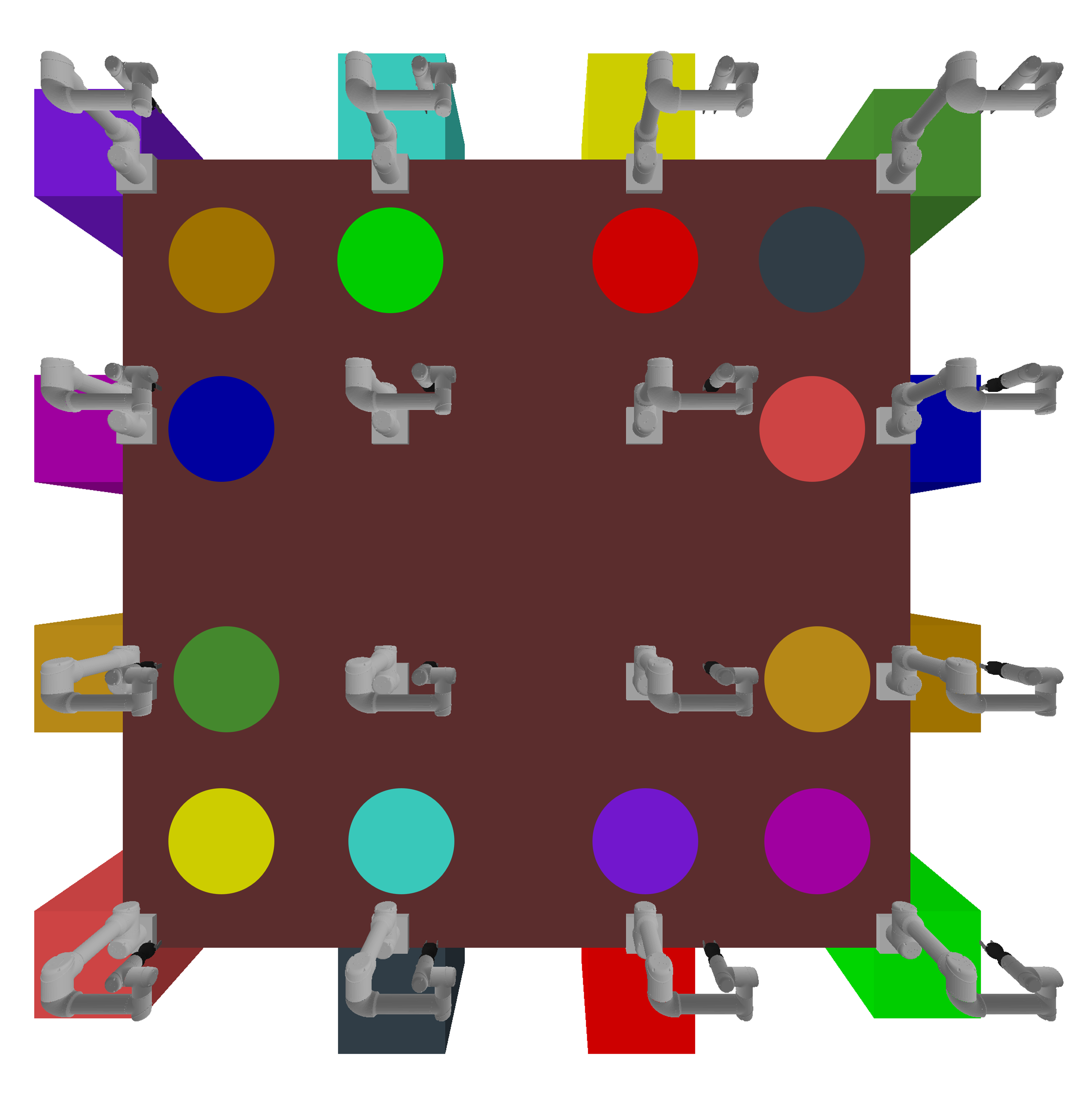}
        \caption{{\color{black}Sorting: 16 robots}}
        \label{fig:16sort}
    \end{subfigure}
    \begin{subfigure}{0.24\textwidth}
        \centering
        \includegraphics[angle=0, width=\linewidth]{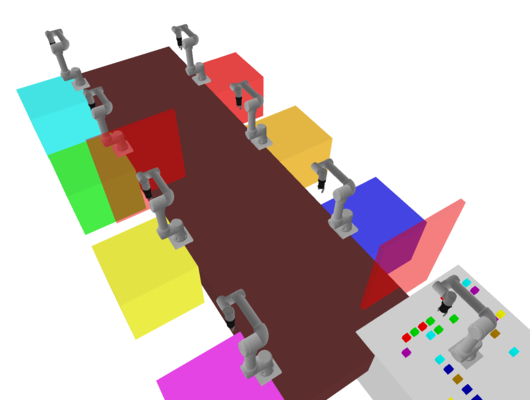}
        \caption{\color{black}Wall: 8 robots / 2 wall}
        \label{fig:1wall}
    \end{subfigure}
    \begin{subfigure}{0.24\textwidth}
        \centering
        \includegraphics[angle=0, width=\linewidth]{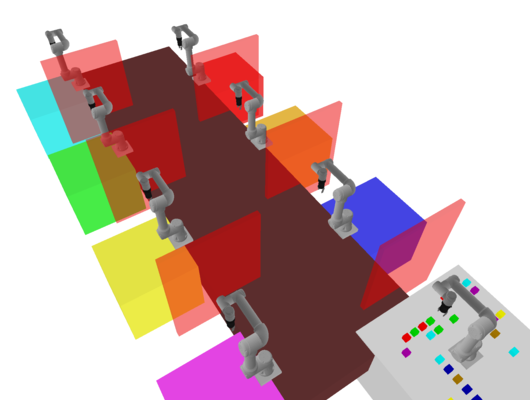}
        \caption{\color{black}Wall: 8 robots / 6 walls}
        \label{fig:3wall}
    \end{subfigure}
    \begin{subfigure}{0.24\textwidth}
        \centering
        \includegraphics[width=0.8\linewidth, trim=2cm 1.5cm 2cm 0cm]{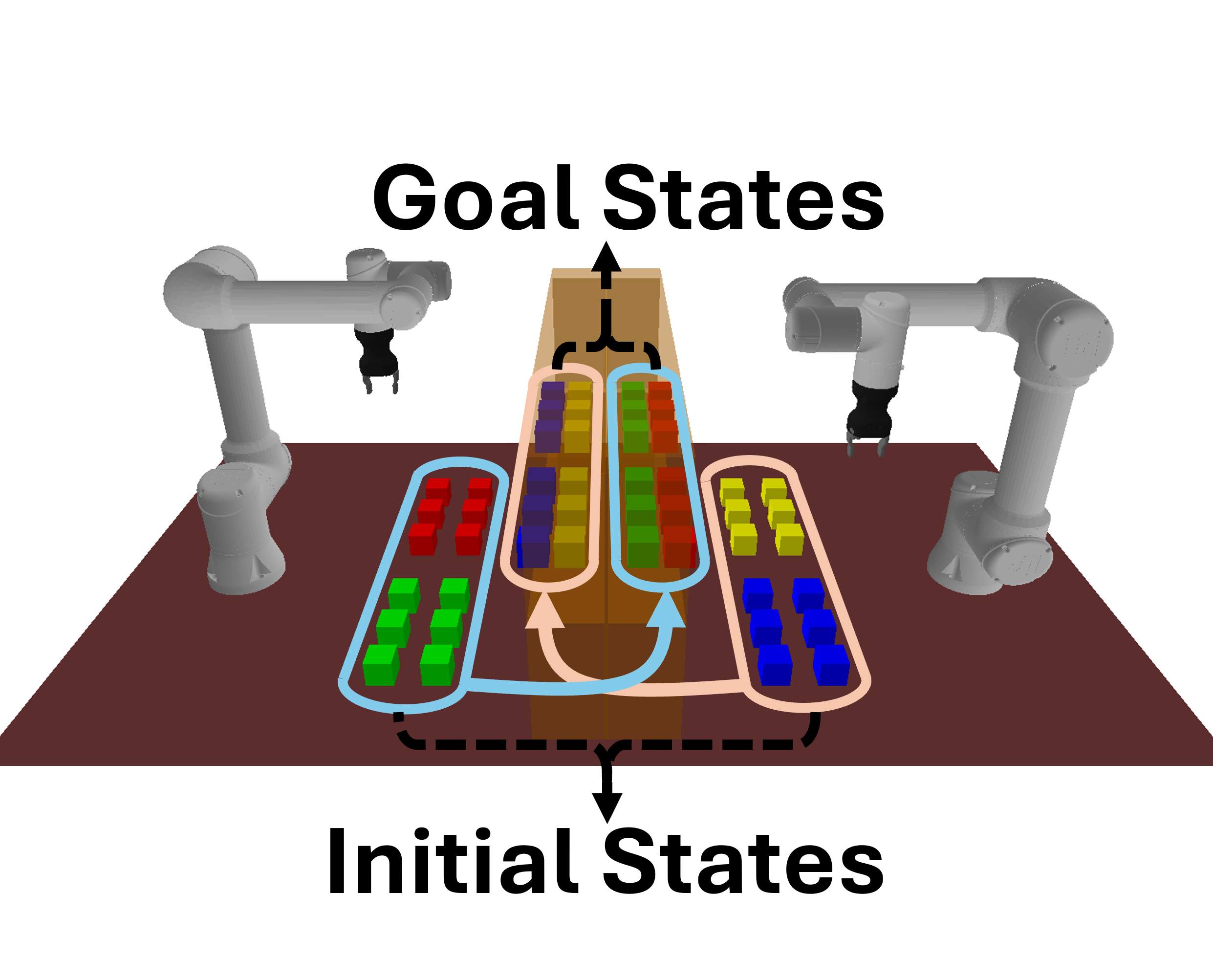}
        \caption{Shelf-wall: 2 robots / 2 shelves}
        \label{fig:shelf-wall2}
    \end{subfigure}
    \begin{subfigure}{0.24\textwidth}
        \centering
        \includegraphics[width=1.0\linewidth]{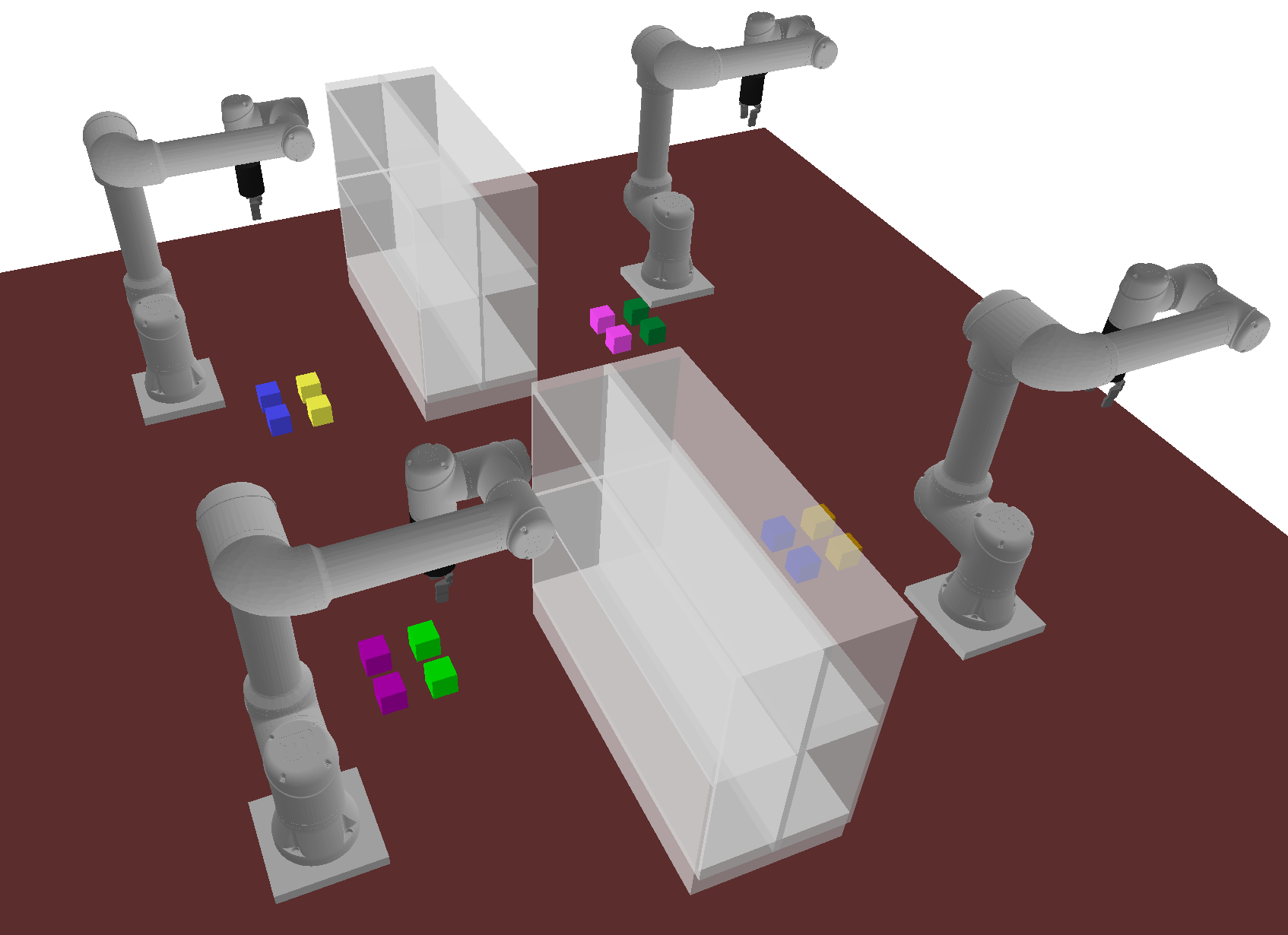}
        \caption{Shelf-wall: 4 robots / 4 shelves}
        \label{fig:shelf-wall4}
    \end{subfigure}
    \begin{subfigure}{0.24\textwidth}
        \centering
        \includegraphics[width=\linewidth]{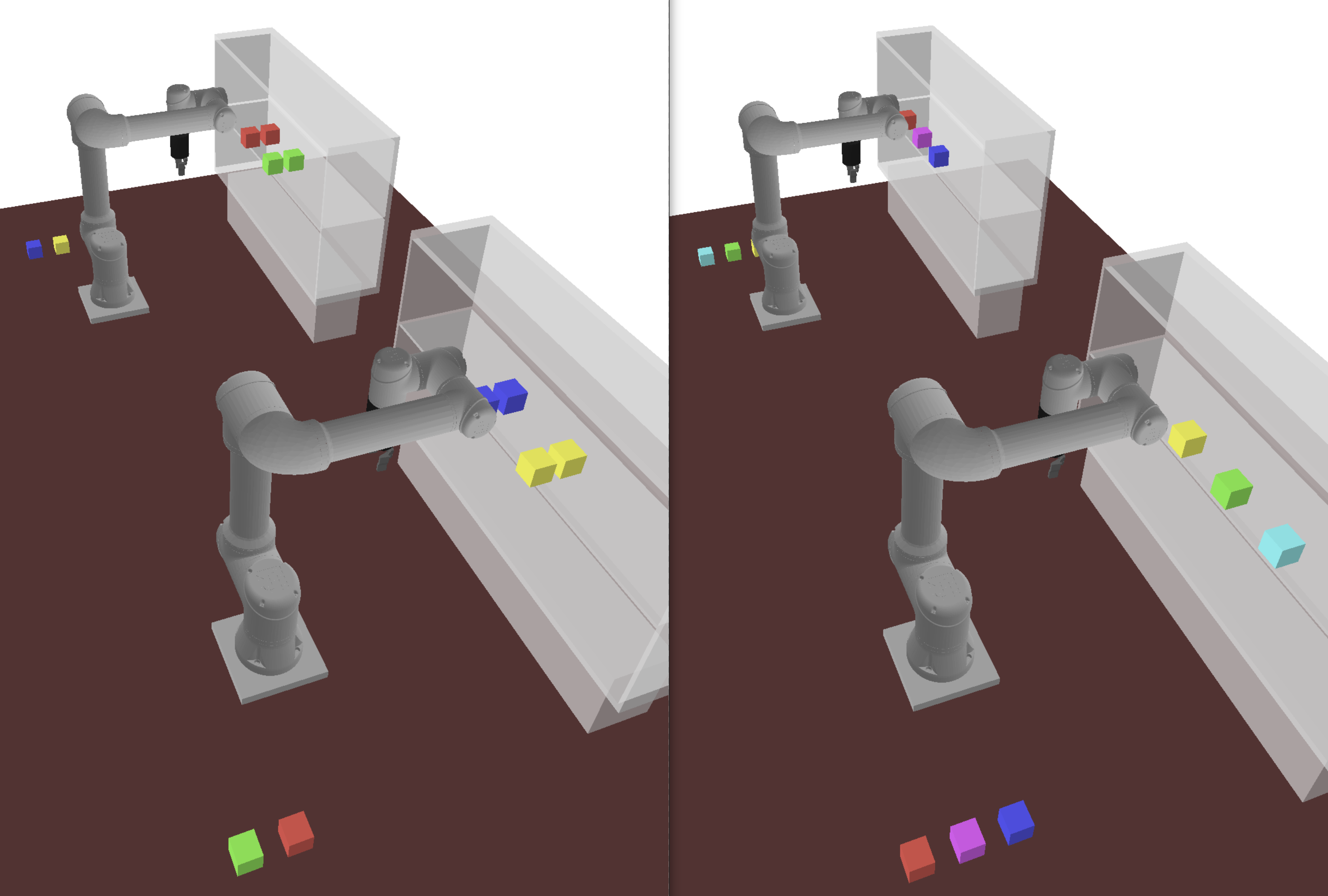}
        \caption{\color{black}Stocking: 2 robots / 3 and 2 layers}
        \label{fig:stocking}
    \end{subfigure}
    \begin{subfigure}{0.24\textwidth}
        % \vspace{10px}
        \centering
        \includegraphics[angle=0, width=\linewidth]{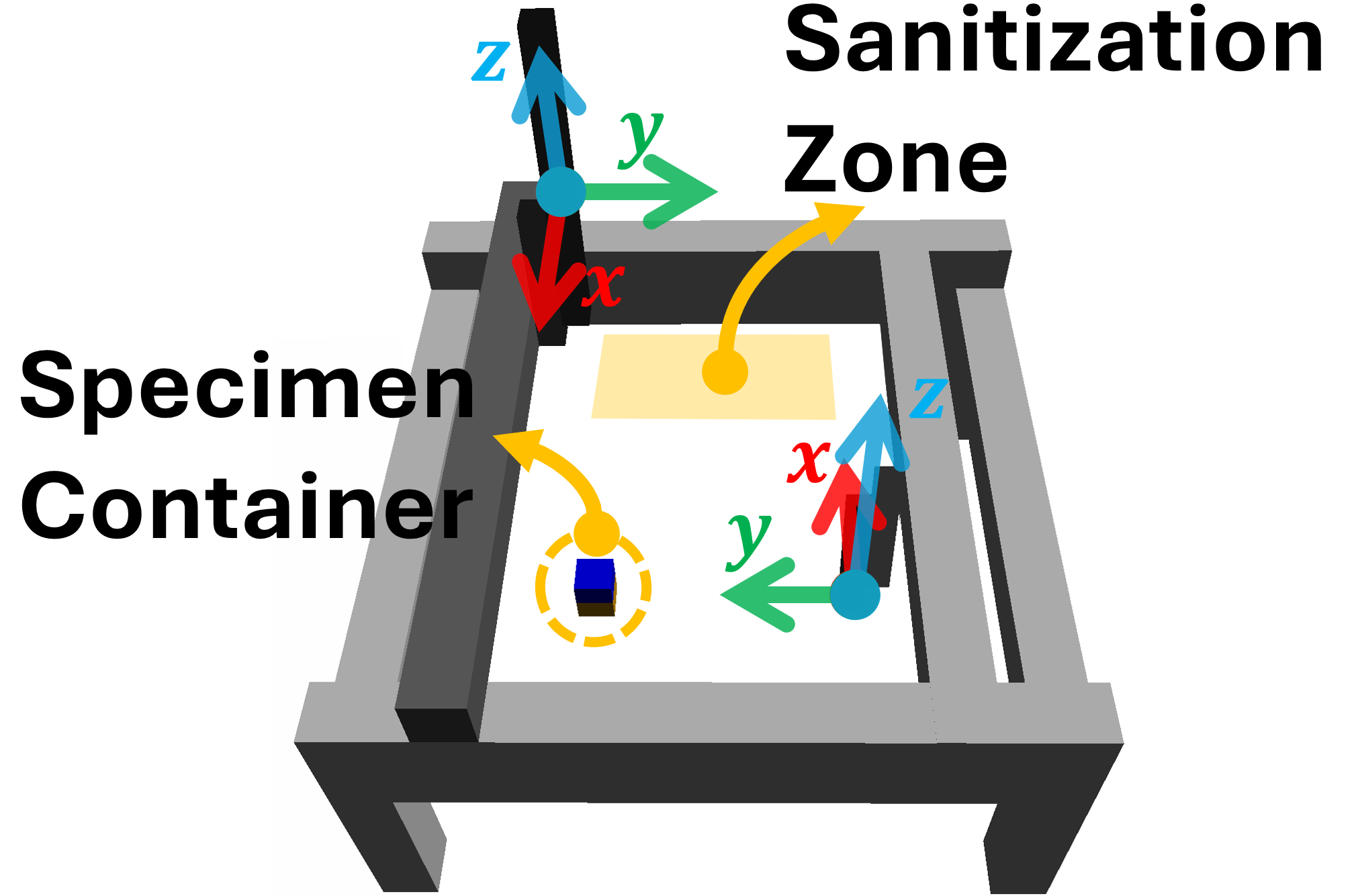}
        \caption{Lab}
        \label{fig:lab}
    \end{subfigure}

    \caption{{\color{black}Five} different types of experiment scenarios. (a) and (b) show ``Sorting" scenarios where the initial clusters of objects are represented within colored circles, and each group must be moved to the matching square boxes. (c) and (d) represent ``Wall" scenarios, featuring different numbers of walls. (e) and (f) illustrate the ``Shelf-wall" scenario, showing the start and goal locations of the blocks. Finally, (g) is the ``Lab" scenario, involving 3-axis gantry robots along with descriptions of the problem entities.}
    \label{fig:scenarios}
\end{figure*}

\begin{figure}
    \centering
    \includegraphics[width=.9\linewidth]{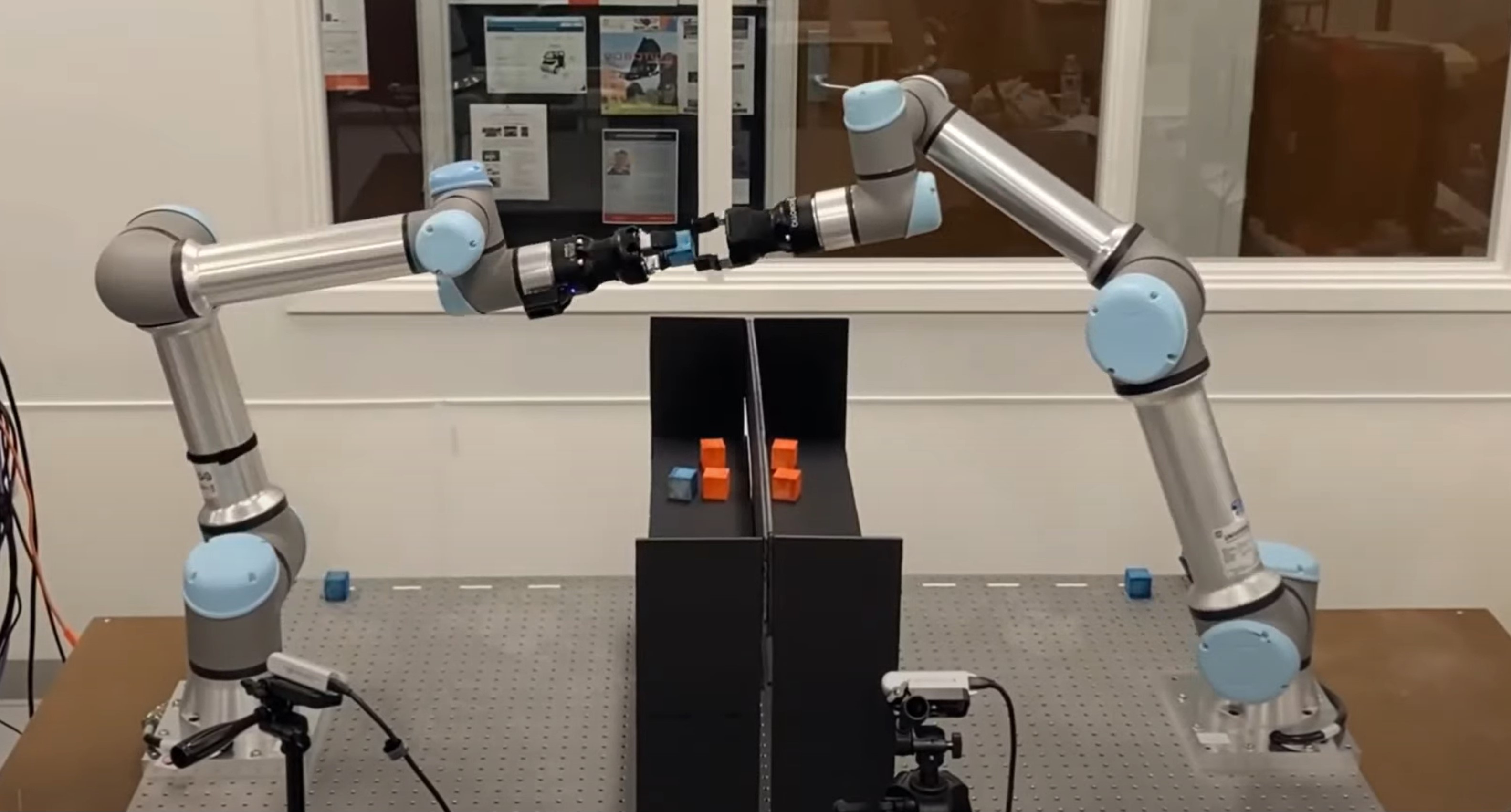}
    \caption{A hardware experiment for the Shelf-wall scenario. The top panels of the shelves have been removed for better visibility. }
    % A full video of demonstration can be found at: \href{https://youtu.be/3eHOzTikcXc}{https://youtu.be/3eHOzTikcXc}}
    \label{fig:hardware-experiment}
\end{figure}

% \begin{figure}
%     \centering
%     \rotatebox{0}{\scalebox{1}[1]{\includegraphics[trim=0.cm 0cm 0cm 0cm, clip, scale=0.6]{figures/sizes_v11.png}}}
%     \caption{Representation size comparison in terms of the number of vertex and hyperarc for Sort, Wall, Shelf-wall, and Lab scenarios. The y-axis represents the number of vertex/hyperarc on a logarithmic scale, while the x-axis denotes the number of objects. }
%     \label{fig:representation_size}
% \end{figure}
\begin{figure*}
    \centering
    \rotatebox{0}{\scalebox{1}[1]{\includegraphics[trim=0.cm 0cm 0cm 0cm, clip, scale=0.33]{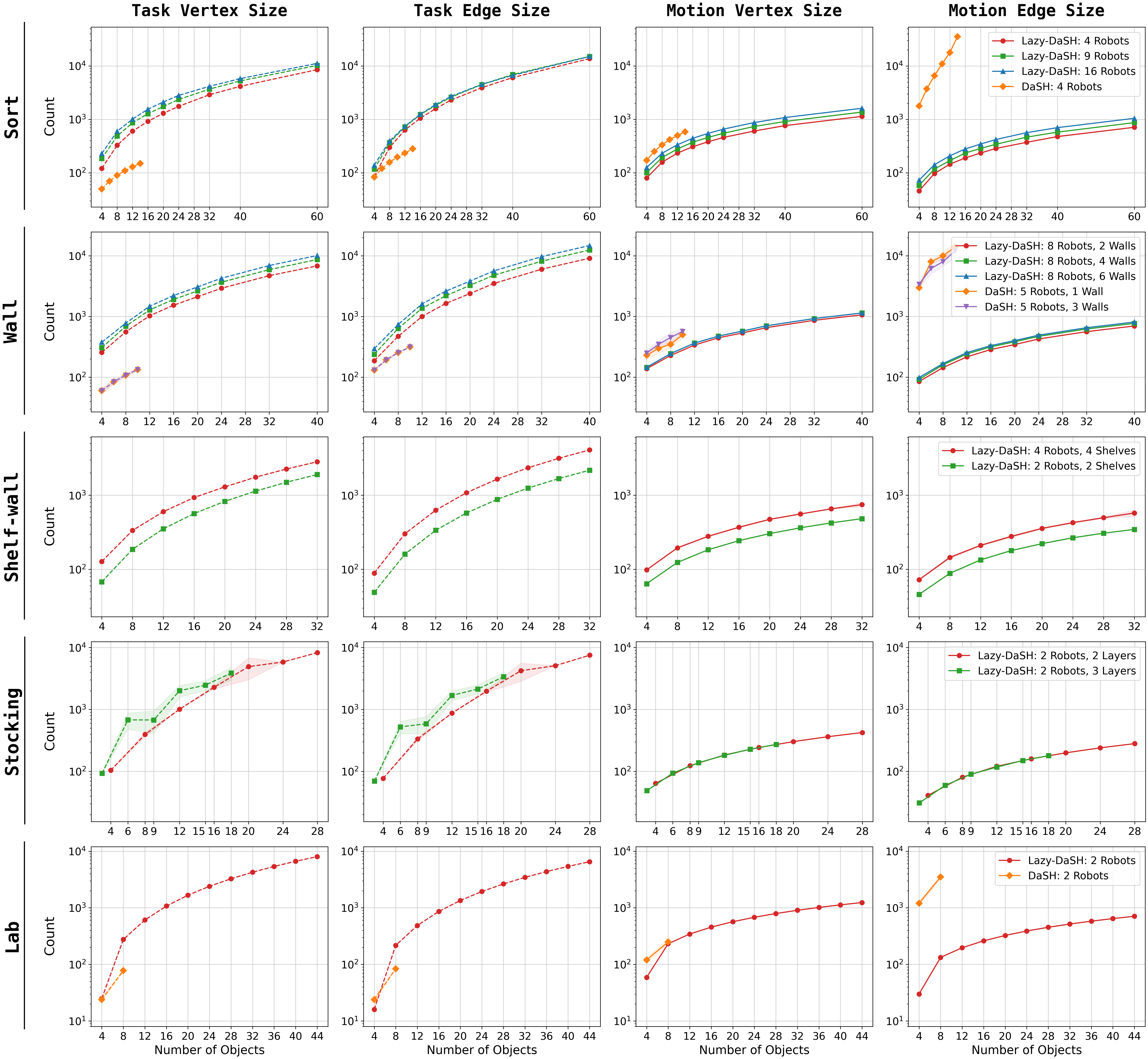}}}
    \caption{{\color{black}Comparison of task and motion representation sizes between Lazy-DaSH and the original framework, DaSH,} measured by the number of vertices and hyperarcs in the Sorting, Wall, Shelf-wall, {\color{black}Stocking}, and Lab scenarios. The y-axis shows the number of vertices/hyperarcs on a logarithmic scale, and the x-axis indicates the number of objects.}
    \label{fig:representation_size}
\end{figure*}

% \begin{figure*}
%     \centering
%     \scalebox{1}[1.0]{\includegraphics[trim=0cm 0.01cm 0cm 0cm, clip,scale=0.7]{figures/planning_time_v11.png}}
%        \caption{A comparison of motion representation construction time (1), task and motion query time (2), conflict resolution time (3), and total planning time (1+2+3) is presented for the Sort, Wall, Shelf, and Lab scenarios. In DaSH, the query time reflects the combined task and motion planning time, whereas in Lazy-DaSH, the query time is the sum of the task query and motion query times. The y-axis represents planning time on a logarithmic scale and is kept consistent across each scenario to clearly compare the proportion of each time component (1, 2, and 3) in the total planning time (1+2+3).}
%     \label{fig:planning_time}
% \end{figure*}
\begin{figure*}
    \centering
    \scalebox{1}[1.0]{\includegraphics[trim=0cm 0.01cm 0cm 0cm, clip,scale=0.33]{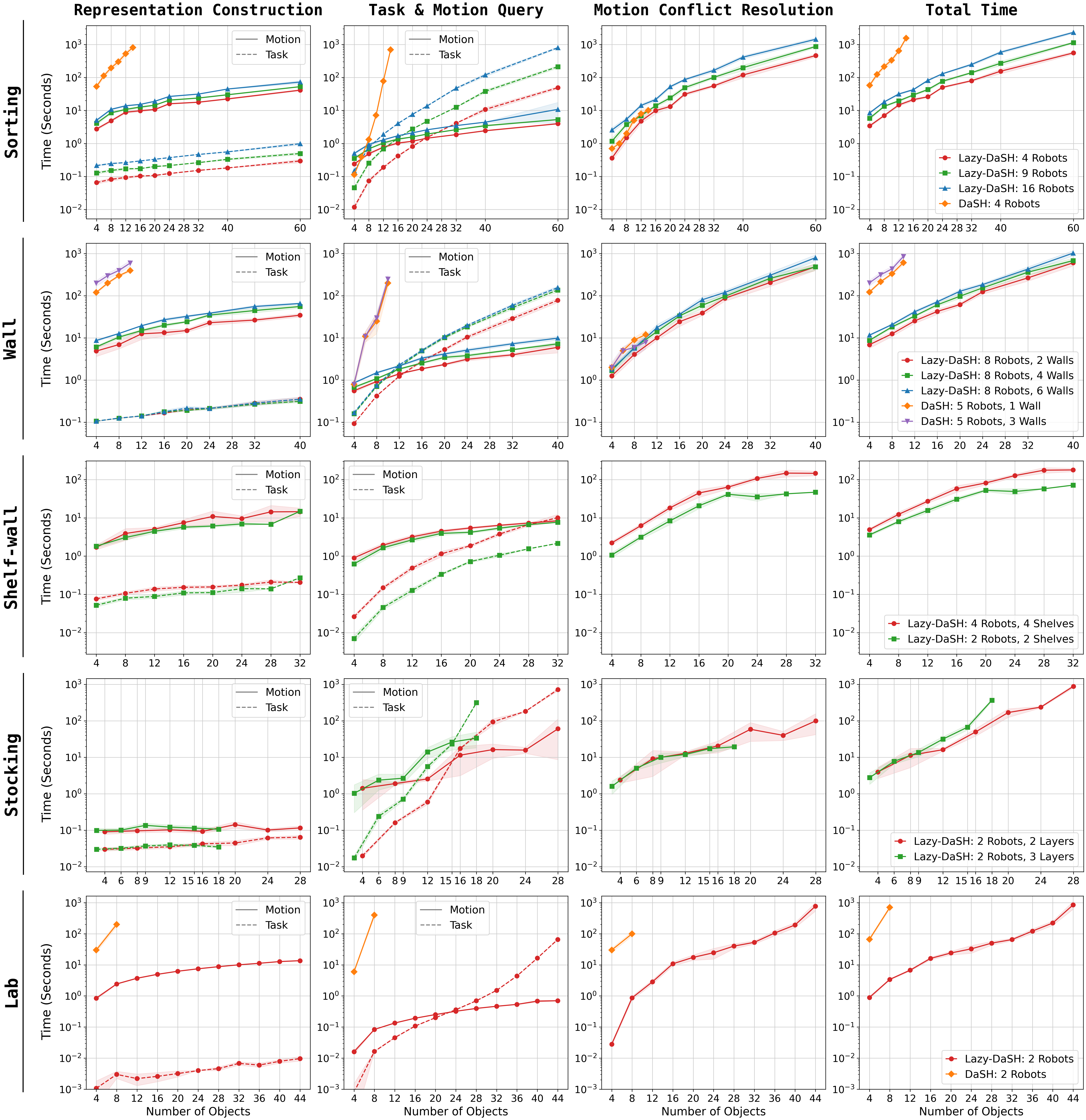}}
       \caption{A comparison of motion representation construction time (1), task and motion query time (2), conflict resolution time (3), and total planning time (1+2+3) is presented for the Sorting, Wall, Shelf-wall, {\color{black}Stocking,} and Lab scenarios. 
       % {\color{orange}(Old) In DaSH, the query time reflects the combined task and motion planning time, whereas in Lazy-DaSH, the query time is the sum of the task query and motion query times.} 
       {\color{black} In DaSH, representation time corresponds to motion construction and query time to the combined task–motion planning. In Lazy-DaSH, by contrast, task- and motion-level times are separated, with dashed lines indicating task and solid lines indicating motion.}
       The y-axis represents planning time on a logarithmic scale and is kept consistent across each scenario to clearly compare the proportion of each time component (1, 2, and 3) in the total planning time (1+2+3).}
    \label{fig:planning_time}
\end{figure*}

\section{\color{black}Validation}\label{sec:validation}
This section details the validation of Lazy-DaSH. 
% {\color{black}(\textbf{Comment:} Can you provide an overview of the result here and also summary of take home?)}
We evaluated Lazy-DaSH across {\color{black} five} scenarios, demonstrating its scalability and efficient constraint management across the hierarchical structure. 
{\color{black} Lazy-DaSH achieves a more compact representation and significantly faster total planning times, demonstrating superior scalability with twice the number of robots and objects than DaSH \cite{motes2023hypergraph}. 
Notably, Lazy-DaSH achieves planning times that are up to an order of magnitude faster than the original framework, which itself has already demonstrated up to three orders of magnitude improvement over the coupled and synchronous planner, Synchronized Multi-Arm Rearrangement (SMART) ~\cite{sb-smrgbmgwcc-20}.}

We begin by outlining the evaluation criteria, experiment scenarios, and method descriptions, followed by an analysis and discussion of the results.
As part of the evaluation, we also demonstrate a hardware experiment for one of the scenarios as shown in Fig.~\ref{fig:hardware-experiment}.

\subsection{Evaluation Criteria}\label{sec:criteria}
{\color{black}
As discussed in Section~\ref{sec:overview}, both Lazy-DaSH and DaSH utilize a hypergraph-based representation to efficiently capture the multi-manipulator object rearrangement problem.
This efficiency arises from the hybrid approach, in contrast to graph-based methods that represent the composite state space.
The hypergraph representation encodes the problem compactly, and querying over this hybrid representation enables faster performance compared to composite state space representations.
Compared to DaSH, Lazy-DaSH preserves this advantage while further improving representation management through its constraint feedback mechanism.

The benefit of the hybrid approach enabled by the hypergraph-based representation has already been validated in the original DaSH work~\cite{motes2023hypergraph} where it demonstrated up to three orders of magnitude speed up in planning times over state-of-the-art composite approaches.
Here we compare directly against DaSH to evaluate the impact of the contributions presented in this paper.
We show that Lazy-DaSH further enhances the hybrid representation by achieving more compact representations and faster query times in complex scenarios.
}
% As discussed in Section~\ref{sec:method_discussion}, Lazy-DaSH maintains a compact yet sufficient representation to ensure successful queries. This results in a smaller representation size compared to DaSH, leading to faster queries and improved scalability. 
% {\color{orange} (Old) To assess these advantages, we evaluate Lazy-DaSH in four distinct scenarios that highlight its capabilities, measuring representation size and overall planning time as the number of robots and objects increases, as shown in Figs.~\ref{fig:representation_size} and \ref{fig:planning_time}.}
{\color{black} To assess these advantages, we measure both representation size and overall planning time as the complexity of each problem increases, {\color{black}comparing results with DaSH} (Figs.~\ref{fig:representation_size} and \ref{fig:planning_time}). 
A detailed analysis is provided in Section~\ref{sec:analysis_and_discussion}.}

\subsection{Method Descriptions}
In this section, we provide implementation comparison of Lazy-DaSH and DaSH.

% {\color{orange} (Old) \subsubsection{DaSH} Probabilistic Roadmap (PRM) \cite{kslo-prpp-96} is used as a motion planner and motion feasibility is all validated during the $\HG_\M$ construction phase. 
% This validation eliminates the need for further motion validation during both the query phase and the conflict resolution phase. 
% A variant of CBS-MP is employed as a motion scheduler to sequence configurations over time in the conflict resolution layer. 

% \subsubsection{Lazy-DaSH} Lazy-Probabilistic Roadmap (Lazy-PRM) \cite{bk-ppulp-00} is used as the motion planner during the $\HG_\M$ construction stage, with motion feasibility evaluated during the motion query and conflict resolution phases. 
% The motion scheduler in the conflict resolution phase is identical to the one used in the DaSH method.}

{\color{black} 
For motion planner, DaSH employs the Probabilistic Roadmap (PRM) method \cite{kslo-prpp-96}, whereas Lazy-DaSH employs the Lazy Probabilistic Roadmap (Lazy-PRM) method \cite{bk-ppulp-00}. 
For a fair comparison of the query process, both methods use the scheduled adaptive robot coordination strategy introduced in Section~\ref{sec:conflict_resolution_layer} for motion conflict resolution, but they differ in their feedback trigger criteria. 
DaSH follows the CBS-MP framework with probabilistic break rules \cite{solis2021representation}, which determine whether to continue expanding the constraint tree or to expand the roadmaps before restarting the query.
}
Both Lazy-DaSH and DaSH were implemented in C++, and the experiments were conducted on a desktop computer equipped with an Intel Core i9-14900K CPU at 3.2 GHz and 64 GB of RAM.

\subsection{Scenarios}\label{sec:scenarios}
Lazy-DaSH and DaSH are evaluated on the multi-manipulator object rearrangement problem, where manipulators are tasked to transport blocks from the start state to the goal state by performing grasp and hand-over actions.
The cube-shaped blocks have randomly generated start and goal positions, ensuring that at least one robot can grasp them and transfer them to the goal. 
{\color{black}For grasp poses, we consider the sides, top, and bottom as possible options.}

% {\color{orange} (Old) We demonstrate four key capabilities: 1) efficient planning for large numbers of robots, 2) an effective constraint feedback mechanism for identifying infeasible robot interactions, 3) resolving geometric constraints, and 4) the ability to solve problems that require more than one step for task completion.}

{\color{black} We demonstrate five key capabilities: (1) scalable planning for large-scale multi-robot systems, (2) an effective constraint feedback mechanism for identifying infeasible robot interactions, (3) resolution of geometric constraints, (4) expansion of the task search space to handle non-monotonicity, and (5) the ability to solve tasks requiring multiple steps for completion.}

To evaluate these capabilities, we designed {\color{black}five} scenarios: \textit{Sorting}, \textit{Wall}, \textit{Shelf-wall}, {\color{black}\textit{Stocking}}, and \textit{Lab}, as illustrated in Fig.~\ref{fig:scenarios}. 
% {\color{orange} (Old) In each scenario, we progressively increase task complexity by adding more robots, resulting in a greater number of interactions needed to transfer objects to their goal locations.
% Additionally, the number of objects gradually increases, further intensifying the task complexity.} 
{\color{black} In each scenario, we progressively increase task complexity based on the features we aim to demonstrate.
In all cases, task complexity grows with the number of objects, and the robots must coordinate and interact to complete the tasks.}
These increments expand the size of both the task space hypergraph and the motion hypergraph, significantly increasing the computational complexity of the query process.

All manipulators in the \textit{Sorting}, \textit{Wall}, \textit{Shelf-wall}, and \textit{\color{black}Stocking} scenarios are UR5e arms equipped with Hand-e grippers, and the \textit{Lab} scenario uses gantry robots with customized end effectors tailored to task requirements. 
We demonstrate a hardware experiment for our \textit{Shelf-wall} scenario, which represents a more complex problem compared to the \textit{Shelf} experiment demonstrated in DaSH \cite{motes2023hypergraph}. 
The corresponding video is provided in the link shown in Fig.\ref{fig:hardware-experiment}.

The tasks for each scenario are described below, along with their design objectives and a summary of the experimental results.

\subsubsection{Sorting}
To assess Lazy-DaSH's efficiency in large-scale planning, we designed Sorting scenarios where four to {\color{black}sixteen} manipulators collaborate to transport colored objects to designated boxes as shown in Figs.~\ref{fig:4sort} and \ref{fig:16sort}.
This is essentially the same as the ``cross sorting" case described in DaSH paper \cite{motes2023hypergraph}, where objects must be delivered to the opposite side of the workstation.
As the number of robots increases, this setup requires more complex coordination of task and motion, as each object must be handed over multiple times among the manipulators.
It is important to note that the minimum number of required interaction grows as the number of robots increases for transfering each object to the goal location.
The results show that Lazy-DaSH significantly improves scalability, successfully handling {\color{black}16} robots with {\color{black}more than 60} objects, whereas DaSH could only handle up to 4 robots with 14 objects.

\subsubsection{Wall}
{\color{black}
This scenario is designed to evaluate how the planner adapts when its optimistic motion assumptions fail due to infeasible robot interactions.
To create such conditions, we introduce thin walls into the Sorting setup, adding obstacles that invalidate the initial lazy assumptions.
As illustrated in Figs.~\ref{fig:1wall} and \ref{fig:3wall}, eight manipulators must transport objects to designated boxes across these walls.
To systematically increase complexity, we progressively add more walls.
Robots positioned near a wall not only sort their own objects but also act as intermediaries, passing items across the obstructions to teammates on the other side.

The result shows that Lazy-DaSH scales up to 8 robots with more than 40 objects, while DaSH scales with up to 5 robots with 10 objects in a presence of interaction obstructions.}

\subsubsection{Shelf-wall}
To demonstrate its ability to handle geometric constraints, DaSH \cite{motes2023hypergraph} introduces the ``Shelf" scenario, where objects must be placed on a shelf such that each rear object is completely blocked by the front object. 
We extend this concept with a more complex scenario, Shelf-wall, which integrates the Shelf and Wall environments, treating each shelf as a barrier that obstructs both manipulator interactions and object access.
As illustrated in Figs.~\ref{fig:shelf-wall2} and \ref{fig:shelf-wall4}, two or four robots are tasked with rearrange objects while satisfying the geometric constraints. 
For example, green (yellow) blocks must be positioned behind the red (blue) blocks and placed on the shelf facing the opposite side. 
Manipulators must hand over objects to robots on the opposite side, as the shelves act as a physical wall between them. 
As the number of robots increases, each object transfer requires at least two handovers, similar to the Sorting scenario.

% {\color{black} The geometric constraints are identified by the task conflict detection layer, which reviews the unvalidated schedule and feeds the identified constraints back to the task query phase so that geometric constraints are incorporated into subsequent queries. 
% Specifically, the task conflict detection layer detects manipulator’s grasp configuration (vertex) collides with statically placed object (vertex)--for example, a green (yellow) block intersecting a fixed red (blue) block.}
The results demonstrate that Lazy-DaSH successfully handles {\color{black}4} robots with {\color{black}32} objects, whereas DaSH fails to find a solution within the time limit even for a 2 robots scenario.

{\color{black}
\subsubsection{Stocking}
% To evaluate the task space representation expansion and non-monotonicity capability of Lazy-DaSH, we consider a real-world stocking scenario, in which two robotic arms place new items on a shelf behind existing items (Fig.~\ref{fig:stocking}). 
% The task search space is expected to be extended since not removing the old items makes it impossible for new object to be placed behind.  
% In addition, the new objects must be taken out and placed back in a specific order so that one cannot obstruct another. 
% The new objects to be stocked are initially placed on the far side of the shelf, similar to the Sorting and Shelf-wall scenarios, so that robot handovers are still required.  

To evaluate the task space expansion and non-monotonicity capabilities of Lazy-DaSH, we consider a real-world stocking scenario in which two robotic arms place new items on a shelf behind existing ones (Fig.~\ref{fig:stocking}). 
Since the existing items must remain in place, the task space must expand; otherwise, placing new objects behind them would be impossible. 
Moreover, the new objects must be taken out and restocked in a specific order to prevent obstruction. 
The new items to be stocked are initially placed on the far side of the shelf, similar to the Sorting and Shelf-wall scenarios, so robot handovers are still required.  

We increase complexity by layering the stocked objects, thereby introducing different levels of non-monotonicity and geometric constraints. 
For example, in a two-layer configuration, the task space must expand for the front objects, which need to be temporarily moved out to place new items at the back. 
In a three-layer configuration, the task space expands for the objects in the first two rows, which must be moved out in sequence and then restocked in reverse order.

The experiments demonstrate that Lazy-DaSH successfully handles three layers of non-monotonicity with 2 robots and 18 objects, and two layers with 2 robots and 28 objects, validating both its scalability and its ability to resolve geometric constraints and non-monotonicity, where DaSH fails.
}

\subsubsection{Lab}
To address problems where each object requires multiple operations to reach its goal state, we introduce the Lab scenario, inspired by wet lab specimen inspection scenarios.
In this scenario, two 3-axis gantry robots inspect specimens within a sealed container, as illustrated in Fig.~\ref{fig:lab}. 
The process begins with the robots removing the lid of the container and placing it in a predefined sanitization zone for sterilization. 
They then use tools attached to their end effectors to inspect the specimen. Once the inspection is complete, the lid is returned to the container to preserve the specimen’s environment. 
Due to the limited space in the sanitization zone, the planner must determine a valid placement that prevents lid collisions. 
This placement is constrained by vertex-vertex conflicts, where each vertex represents a potential lid position in the sanitization zone. 
% {\color{black} Any vertex--vertex conflicts are first identified by the task conflict detection layer and are resampled if collisions exist in the unvalidated schedule.}
The results demonstrate that Lazy-DaSH outperforms DaSH, successfully handling more than twice the number of objects requiring multi-stage operations.

\subsection{{\color{black}Analysis and Discussion}}\label{sec:analysis_and_discussion}
{\color{black}This section analyzes and discusses the experiment results, focusing on {\color{black}five} distinct capabilities of Lazy-DaSH using the criteria defined in Section \ref{sec:criteria}.}

\subsubsection{Overview} 
Across all tested scenarios, Lazy-DaSH demonstrated significantly improved scalability compared to DaSH, as illustrated in Figures~\ref{fig:representation_size} and~\ref{fig:planning_time}. 
The scalability analysis in~\cite{motes2023hypergraph} shows that the number of objects leads to linear growth in representation size, while the number of robots results in quadratic growth.
This makes DaSH particularly inefficient in environments with many robots, where the exhaustive expansion of the search space imposes substantial computational overhead.
This limitation was evident in the Sorting scenarios, which are designed to test scalability with respect to the problem size in terms of both the number of robots and objects.

{\color{black}
DaSH also struggles in scenarios that require frequent replanning at both the task and motion levels because it primarily incorporates motion level constraints without adapting the higher level task structure.
This limitation is particularly evident in environments with complex constraints such as the Wall scenario, where Lazy DaSH’s optimistic motion validation produced infeasible motions and required constraint feedback to handle invalid robot interactions during the replanning phases.
Similarly, the Shelf wall scenario, which is strongly constrained by geometric rules, was not solvable by DaSH.
In Lazy DaSH, however, such geometric constraints are identified by the task conflict detection layer, which reviews the unvalidated schedule, detects when a manipulator’s grasp configuration (a vertex) collides with a statically placed object (another vertex), and feeds these constraints back into the task query phase.

DaSH also struggles with non monotonic scenarios because it pre samples object poses without reasoning about their necessity, many of which never contribute to the plan.
In contrast, Lazy DaSH expands task space elements only when required.
For example, in the Stocking scenario, non monotonicity is identified by the task conflict detection layer, which then triggers selective expansion of the task space.
In the Lab scenario, vertex–vertex conflicts (e.g., placing two lids in the same spot) are similarly identified and resolved by resampling to ensure valid placements.

Thus, the hierarchical query framework, combined with a targeted constraint feedback mechanism, enables focused refinements of task and motion representations, thereby improving both efficiency and scalability in complex environments.
}

\subsubsection{Representation Size}
The planning performance of both DaSH and Lazy-DaSH is directly influenced by the size of their representations, which form the foundation for iterative queries. We compare the sizes of {\color{black}$\HG_\T$ and $\HG_\M$} in terms of the number of vertices and hyperarcs, as shown in Figure~\ref{fig:representation_size}.

{\color{black}As shown in the first two columns, the task space representation of Lazy-DaSH is an order of magnitude larger than that of DaSH, due to the inclusion of start and goal object-only task space elements and their associated robot transitions.
In contrast, in comparable scenarios such as Sorting, Wall, and Lab, the motion representation in DaSH, requiring costly collision checking, can be up to two orders of magnitude larger than in Lazy-DaSH.
This makes it difficult for DaSH to query over the motion representation using its combined task and motion query approach.
Lazy-DaSH, on the other hand, benefits from its constraint management system, which selectively expands only the necessary spaces, leading to a more compact and efficient representation.
This difference is particularly evident in the rapid growth of motion hyperarcs in DaSH compared to Lazy-DaSH, especially in the Sorting and Wall scenarios, which involve four and five robots, respectively.
Task space expansion is also observed in the Stocking scenarios, where non-monotonicity triggers growth in the representation; in these cases, the number of vertices and hyperarcs varies as object poses are resampled to avoid collisions.

The effectiveness of this targeted motion representation expansion and lazy motion validation is further evaluated in terms of planning time in the next section, with their impact on overall planning performance analyzed in the subsequent discussion.}% reduction in representation size demonstrates the effectiveness of Lazy-DaSH’s task query phase in identifying only the necessary task space elements.
% The results show that the number of vertices and hyperarcs in $\HG_\M$ is significantly smaller in Lazy-DaSH across all scenarios. 
% Notably, the number of hyperarcs—critical due to their impact on motion validation checks—is up to two orders of magnitude smaller in Sorting, Wall, and Lab, and up to one order of magnitude smaller in Shelf-wall compared to DaSH.

\subsubsection{Planning Time}
Fig.~\ref{fig:planning_time} presents the total planning time along with a detailed breakdown. 
The first three columns represent the key components contributing to the total planning time: {\color{black}task and motion representation construction}, task and motion query, and conflict resolution. 
The fourth column shows the total planning time, which is the sum of these three components.
{\color{black}
Task conflict detection is included in the total runtime, but it is only on the order of milliseconds, since it relies on simple per-configuration collision checks that are far less costly than motion queries.}
To facilitate direct comparisons across different planning times, the y-axis for each scenario is kept consistent across all columns.

{\color{black}The representation construction time for both DaSH and Lazy-DaSH are separately shown in task and motion, showing that the motion representation accounting for the largest portion due to the computational cost of collision checking.}
In DaSH, this process involves constructing a fully validated roadmap upfront. In contrast, Lazy-DaSH adopts a lazy construction approach, initially generating an edge-invalidated roadmap and incrementally refining it in response to failures encountered during motion queries or conflict resolution. 
In addition to the reduced representation size in Lazy-DaSH, this incremental approach significantly decreases construction time, achieving up to two orders of magnitude reduction in scenarios where both methods are comparable.

The query times for DaSH and Lazy-DaSH differ in definition. 
In DaSH, it represents the combined task and motion planning query time, whereas in Lazy-DaSH, {\color{black}it is differentiated with task and motion queries}. 
In both methods, queries may be iteratively triggered by failures in planning or conflict resolution, and the cumulative time for these iterations is reflected in the total query time.
A key observation is that as the number of objects increases, the query time in DaSH grows at a much steeper rate compared to Lazy-DaSH. This is due to DaSH’s representation size becoming intractable, leading to an exponential increase in search space. 
In contrast, Lazy-DaSH maintains a more scalable query process, significantly reducing query time. The reduction reaches up to three orders of magnitude in scenarios where DaSH is able to find a solution. 

{\color{black}Motion conflict resolution, involving iterative detection and resolution of path conflicts, takes a noticeable share of planning time as the number of objects grows.}
% This is inherent to the conflict-based search appoach \cite{solis2021representation}, where motions are recomputed and entire paths are revalidated for collisions as conflicts are detected. 
As the length of the motion plan increases, typically due to an increased number of objects, the likelihood of recomputation and revalidation increases, becoming a primary source of computation in the overall planning process.
Although this suggests that the conflict resolution layer could become a bottleneck in the planning process as the problem size increases, it is worth noting that this layer can be replaced with off the shelf schedulers or their variants to improve scalability. 
% Future work will explore integrating alternative motion scheduling techniques into the conflict resolution layer to further enhance scalability.

\section{Conclusion}
In this paper, we introduce Lazy-\textbf{D}ecomposable St\textbf{a}te \textbf{S}pace \textbf{H}ypergraph (Lazy-DaSH), a novel hypergraph-based approach to multi-robot object rearrangement that extends DaSH.
Lazy-DaSH separates task and motion planning into distinct layers connected through a constraint feedback mechanism and employs a lazy motion evaluation strategy, validating only the motions relevant to the candidate task plan to minimize unnecessary computation.
{\color{black}The constraint feedback mechanism manages infeasible {\color{black}task orders and motions} identified by the dedicated conflict detection layer, enabling dynamic updates and refinements of both task and motion planning representations.
This design allows the planner to maintain a concise representation, support efficient representation-driven queries, and incorporate the constraints critical to task completion.}
Experimental results across {\color{black}five} scenarios demonstrate that Lazy-DaSH significantly improves scalability and planning speed in highly constrained environments compared to DaSH.
Additionally, one of the experiments is validated through hardware implementation:\href{https://youtu.be/3eHOzTikcXc}{https://youtu.be/3eHOzTikcXc}.
{\color{black}
As future work, adapting online execution into the Lazy-DaSH framework to enhance robustness in dynamic environments represents a promising research direction.}
\bibliographystyle{IEEEtran}
\bibliography{main}

\end{document}